\documentclass{article}
\usepackage{spconf,amsmath,graphicx}
\usepackage{url}

\title{Evaluating Automatic Speech Recognition Systems in Comparison with Human Perception Results Using Distinctive Feature Measures}
\name{Xiang Kong$^1$, Jeung-Yoon Choi$^2$, Stefanie Shattuck-Hufnagel$^2$}
\address{$^1$Department of Computer Science, University of Illinois at Urbana-Champaign\\
$^2$Speech Communication Group, Research Laboratory of Electronics, MIT\\
xkong12@illinois.edu, jyechoi@mit.edu, sshuf@mit.edu}
\begin{document}
%
\maketitle
\begin{abstract}
This paper describes methods for evaluating automatic speech recognition (ASR) systems in comparison with human perception results, using measures derived from linguistic distinctive features. Error patterns in terms of manner, place and voicing are presented, along with an examination of confusion matrices via a distinctive-feature-distance metric. These evaluation methods contrast with conventional performance criteria that focus on the phone or word level, and are intended to provide a more detailed profile of ASR system performance, as well as a means for direct comparison with human perception results at the sub-phonemic level.
\end{abstract}
\section{Introduction}
\label{sec:intro}

Conventional automatic speech recognition (ASR) systems model and evaluate the speech recognition process as follows. Acoustic measurements can be expressed as a sequence $X=\{x_1, x_2, ..., x_t, ..., x_T\}$, and the true word sequence as $W=\{w_1,w_2,⋯,w_n,⋯,w_N\}$, so that the optimum estimated sequence is $\widehat{W} = \arg\max P(W|X)$, or more generally, with a scoring function, $S(W,X)=> \widehat{W} = \arg\max P(W|X)$. In order to compare the difference between  $W$ and $\widehat{W}$ , the most commonly used evaluation measure is the word error rate (WER) \cite{he2011word}. The general form of an error rate is given by error rate = (deletions + insertions + substitutions) / (entries), and accuracy is (1 – error rate). For different recognition tasks, other evaluation measures can be applied. For binary classification applications, such as keyword spotting, the F-score may be used, which is calculated as the harmonic mean of the precision and recall. To evaluate the word accuracy of translation from one language to another, the BLEU (Bilingual Evaluation Understudy) method uses the correspondence of the translation between the machine and humans, and requires word boundaries. Similarly, the ROUGE method evaluates word accuracy by comparing the translation from a machine to a human-produced summary \cite{papineni2002bleu} \cite{lin2004rouge}.

A common characteristic of the evaluation methods mentioned above is that the metrics are related to word-level matches. A drawback of evaluation measures at the word level is that they capture overall system performance, combining the results of acoustic and language models. A simple way to isolate the performance at the acoustic level is to evaluate at the phonemic level, using a unit such as the phone, leading to phone error rates (PER), which are generally lower than the WER. However, using PER to describe performance at the acoustic level is still problematic in at least two aspects. First, the phone sequence for a produced word may vary widely, especially for common function words and for word sequences susceptible to reduction, so that direct comparison of pronunciation variants becomes difficult. Second, it is not possible to measure how similar two phones are to each other using phones as analysis units. The first problem may be solved by restricting the test sequences to utterances with no pronunciation variants; an example would be VCV (vowel-consonant-vowel) syllables. The second issue can be addressed by describing phoneme level units at a sub-phonemic level, by using linguistic distinctive features \cite{halle1992phonological}. For example, the phonemes \textbf{/f/} and \textbf{/s/} are both \textbf{[+consonantal]}, \textbf{[-sonorant]}, \textbf{[+continuant]}, but \textbf{/f/} is \textbf{[+labial]}, while \textbf{/s/} is \textbf{[+alveolar]}. 

In line with such research, this paper describes methods for evaluating system performance using distinctive feature measures, for a detailed analysis of results at the sub-phonemic level. This approach enables descriptions of the non-uniform effects on recognition results from perturbations of the input speech, such as ambient noise, or from non-standard speaker characteristics, such as foreign accents. This approach also allows direct comparison with human perception results, and may provide directions for modeling human perception patterns more closely in ASR systems. Accordingly, in this paper, we observe the acoustic-level performance of two types of ASR systems perturbed with additive white noise in comparison with human perception results, using VCV syllables. ASR systems based on Hidden Markov Models (HMMs) and Deep Neural Networks (DNNs) are examined. In contrast to conventional evaluation methods, performance is analyzed by classifying error patterns related to manner, place and voicing, and by examination of confusion matrices via a distinctive-feature-distance metric.

\section{Methods}
\label{sec:Methods}
\subsection{Databases}
The TIMIT database was used for training the ASR systems, and the LAFF VCV database with additive noise was used as the test set. TIMIT \cite{garofolo1993darpa} contains broadband recordings of 630 speakers of 8 major dialects of American English, each reading 10 phonetically rich sentences, and includes time-aligned orthographic, phonetic and word transcriptions, as well as a 16kHz speech waveform file for each utterance. LAFF VCV \cite{laff} is a collection of vowel-consonant-vowel syllables recorded at the MIT Speech Communication Group from 2 male speakers and 1 female speaker who were native speakers of North American English. The syllables were formed from 6 vowels and 26 consonants, resulting in utterances such as \textbf{/aa-b-aa/}, etc. Full-band white noise was added to the VCV utterances, to produce test stimuli with various signal-to-noise ratios.
\subsection{Conversion of Miller \& Nicely results}
First, the results from the confusion matrices in Miller and Nicely’s paper \cite{miller1955analysis} on human perceptual confusions in noise were converted to manner, place and voicing errors as well as grey-scale confusion matrices, for more direct comparison with other results. Manner errors involve errors between vowels, glides = \{w, y, r, l, h\}, nasals = \{m, n, ng\}, fricatives = \{f, th, s, sh, v, dh, z, zh\}, stops = \{p, t, k, b, d, g\}, and affricates = \{ch, dj\}. Place errors indicate errors between labials = \{m, f, v, p, b\}, dentals = \{th, dh\}, alveolars = \{n, s, z, t, d\}, palatals = \{sh, zh, ch, dj\}, and velars = \{ng, k, g\}. Voicing errors are between unvoiced = \{f, th, s, sh, p, t, k\} and voiced = \{all others\}. In the grey-scale confusion matrices shown in Figs. 5 - 8, the magnitude of the entry in each cell corresponds to the darkness of the cell (e.g. 10 out of possible 10 trials would be 100\%, or black). The additive white noise levels range from 12dB SNR to -18dB SNR, in 6 dB SNR decrements. In the following results, the entries in the confusion matrices are in the order of \textbf{/p, t, k, f, th, s, sh, ch, b, d, g, v, dh, z, zh, dj, m, n, ng, w, y, r, l, h/}, and “No Response,” with entries skipped if not present. Unvoiced sounds appear before voiced sounds, with stops preceding fricatives, and nasals and glides are at the end. This placement facilitates the visual analysis of error patterns in the grey-scale confusion matrices.

\subsection{Human perception of LAFF VCV in noise}
Next, human listening experiments were carried out with selected LAFF VCV data. The stimuli consisted of the following six syllables: \textbf{/aa-b-aa/}, \textbf{/aa-d-aa/}, \textbf{/aa-s-aa/}, \textbf{/aa-m-aa/}, \textbf{/aa-ch-aa/}, and \textbf{/aa-sh-aa/}. Each syllable was embedded in full-band white noise ranging from 30dB SNR to -20dB SNR in 10dB decrements. Twenty adults (17 females, 3 males) between 18 and 31 years of age (mean = 22, SD = 3) were recruited for participation in the experiment. All participants were monolingual speakers of American English and had no history of speech, language, hearing, or neurological disorders according to self-report.  Each participant heard 360 syllables (6 consonants $\times$  10 repetitions $\times$  6 SNRs). The syllables were blocked by SNR, with block order randomly determined for each participant, and syllable order randomized within each block. Participants were directed to provide a response even when they were not completely sure of the item. Stimulus presentation and data collection was controlled using the SuperLab software \cite{cedrus}. 
\subsection{Hidden Markov Model(HMM)-based ASR system}
As an example of a widely-used HMM-based ASR system, we selected the standard HTK system \cite{young2002htk}. Standard training methods for the HTK system were applied to obtain single-phone models (6 states with non-emitting first and last states/8-mixture Gaussian mixture models), and HInit and HRest procedures were used to initialize models before running HERest for model optimization \cite{woodland1993htk}. The results showed 67\% correct phone recognition for the entire TIMIT test set, and 65\% for the TIMIT core test set. Error rates reported in the literature similarly ranges from 46\% to 55\%, in line with our error rates. For our study, we focussedd on detection of consonants and glides, selecting 36 speech files from the LAFF database, each of which included a consonant or glide between two "\textbf{aa}" phones, such as \textbf{/aa-z-aa/}, \textbf{/aa-b-aa/}, etc. White noise at various levels from 40dB SNR to -20dB SNR at 10dB decrements were added to form the test data set \cite{hawkins1950masking}. (A subset of these files was selected for use in the human perception tests in Section 22.5.) Acoustic feature files were obtained using Hcopy, the consonants were detected using HVite. The HTK results were matched to the true sequence (minimum edit distance match), and the results were tabulated in terms of error types and into grey-scale confusion matrices.
\subsection{Deep Neural Network(DNN)-based ASR system}
For the DNN training \cite{hinton2012deep}, 24 Mel-scale log filter banks are extracted as input features. There are five hidden layers, each of which has 2048 nodes, and additionally there is a softmax output layer; we use backpropagation to tune the weights. Overall, the DNN system was implemented using the KALDI toolbox \cite{povey2011kaldi}, and trained on the TIMIT TRAIN set, resulting in a PER of 30.87\% on the TIMIT TEST set.

\begin{figure}[!htb]
\begin{minipage}[t]{0.45\linewidth}
    \centering
    \includegraphics[width=1\textwidth]{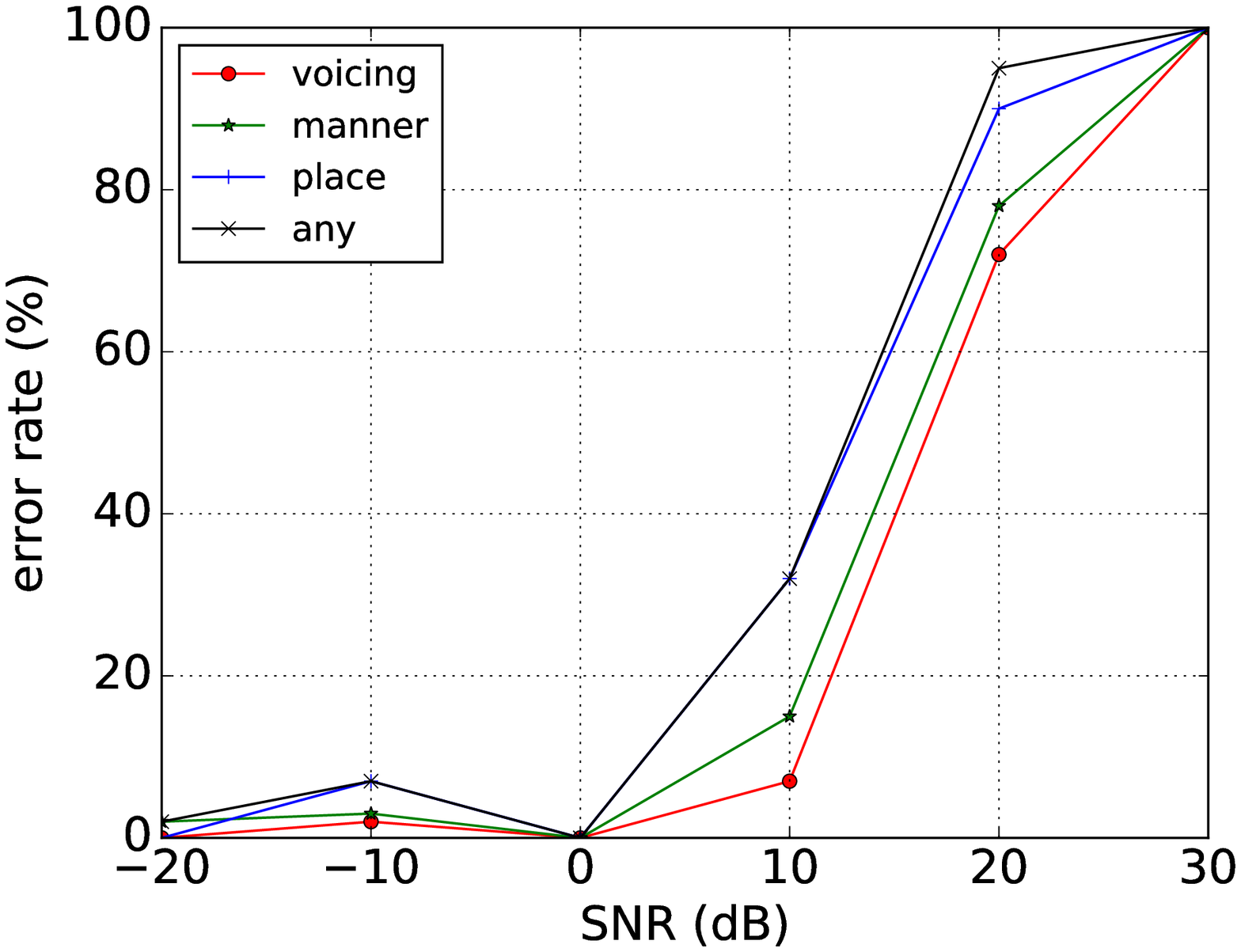}
    \caption{Error patterns for human perception results}
    \label{humanper}
\end{minipage}
\hspace{0.1cm}
\begin{minipage}[t]{0.45\linewidth} 
    \centering
    \includegraphics[width=1\textwidth]{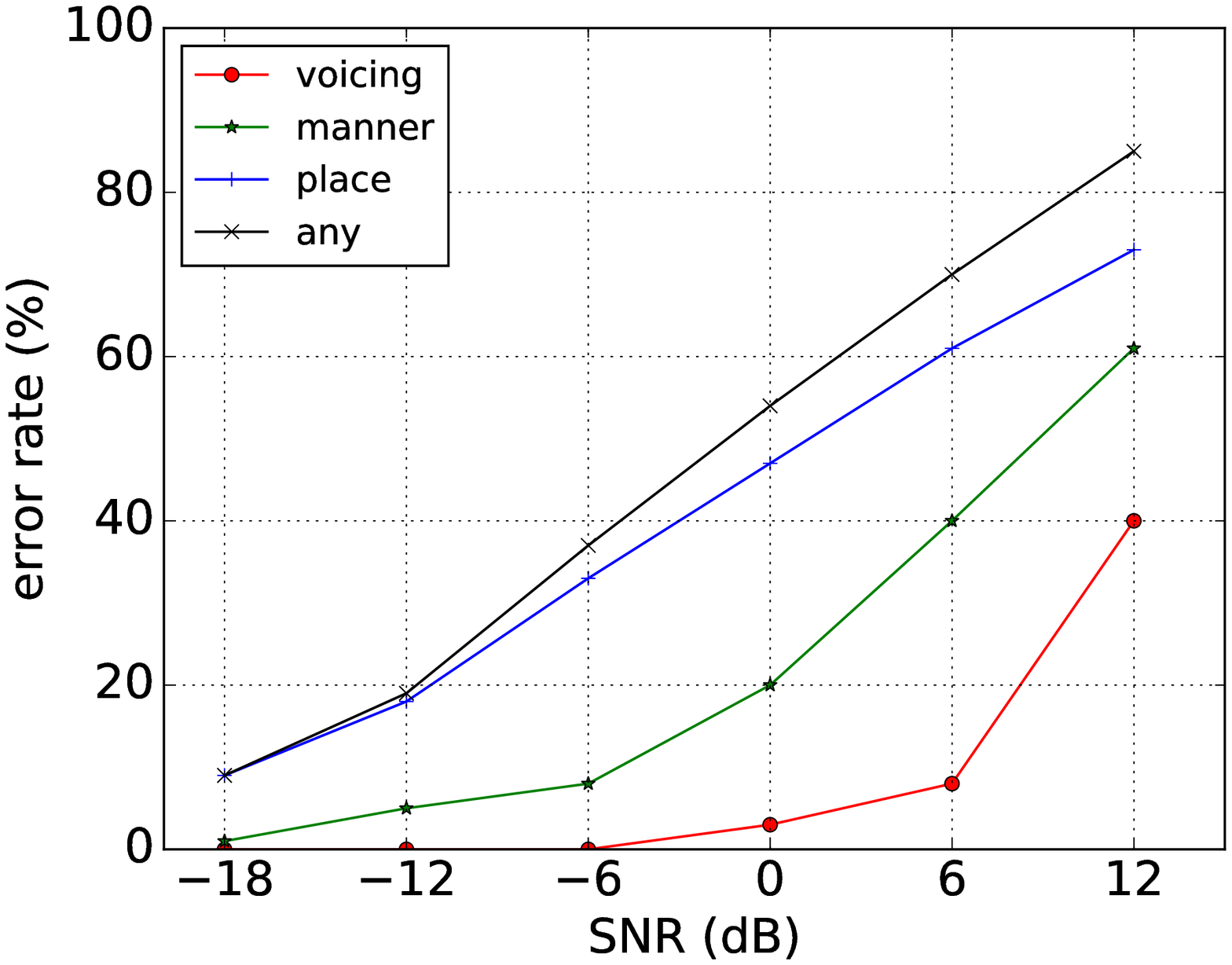}
    \caption{Error patterns for Miller \& Nicely results}
    \label{miller}
\end{minipage}        
\end{figure} 

\begin{figure}[!htb]
\begin{minipage}[t]{0.45\linewidth}
    \centering
    \includegraphics[width=1\textwidth]{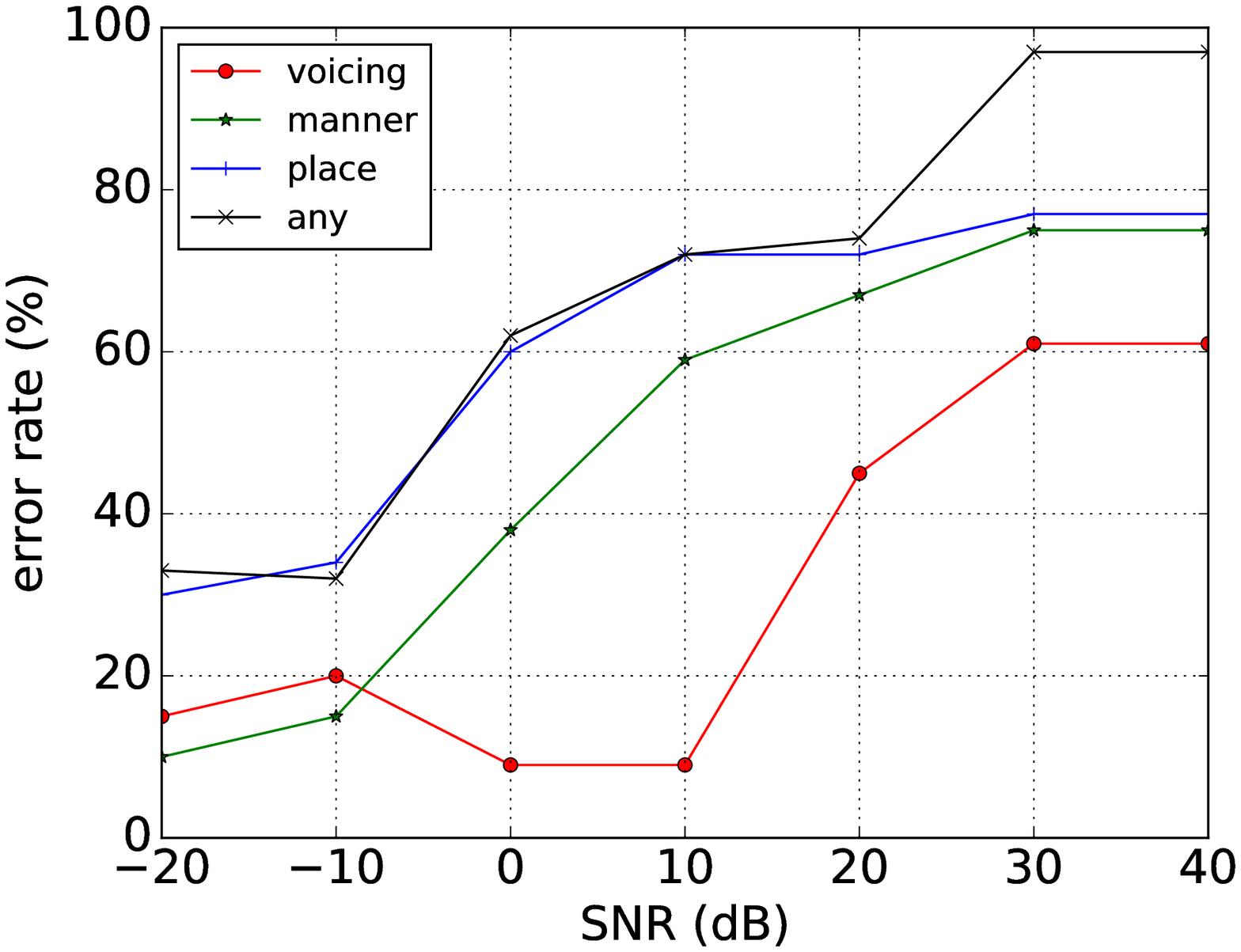}
    \caption{Error patterns for HMM-based system results}
    \label{hmm}
\end{minipage}
\hspace{0.1cm}
\begin{minipage}[t]{0.45\linewidth} 
    \centering
    \includegraphics[width=1\textwidth]{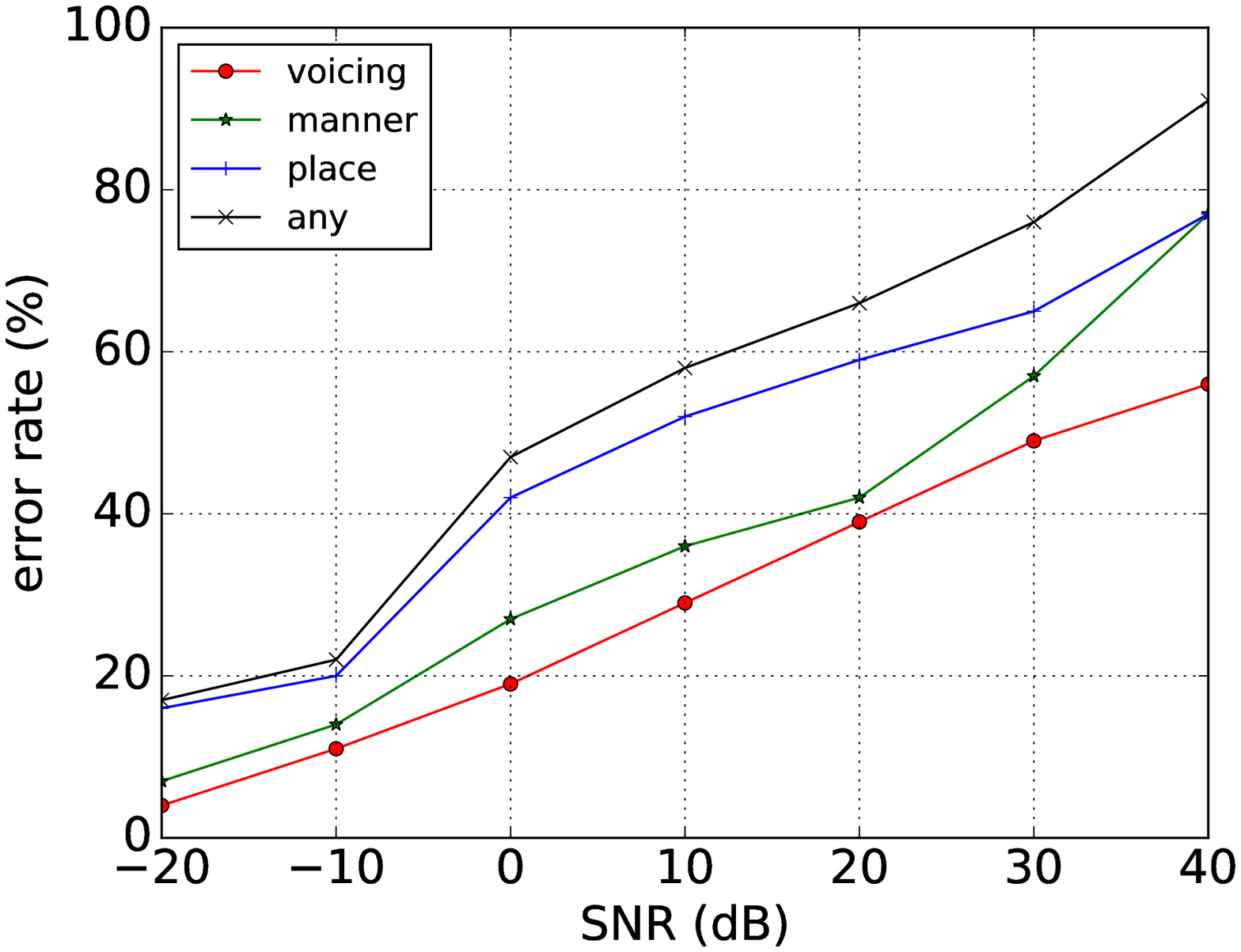}
    \caption{Error patterns for DNN-based system results}
    \label{dnn}
\end{minipage}        
\end{figure} 

\section{Results}
\label{sec:results}
\subsection{Manner Place and Voicing error patterns}
Human perception results for LAFF VCV database files are shown in Fig. 1. The results show that as SNR decreases, error rates increase, particularly after 0 dB SNR. Also, place errors are most prevalent, followed by errors in manner, then voicing. Next, the results converted from the Miller and Nicely paper are shown in Fig. 2. Similar to the results for the LAFF VCV database, errors increase monotonically, with place errors highest, followed by manner errors, then voicing errors. In comparison, results for HMM phone detection in the LAFF VCV database are shown in Fig. 3. In contrast with the results for human perception, it is not always the case that error rates increase monotonically as SNR decreases. Also, the relative degree of errors in place, manner and voicing are not consistent as in the human perception results. Nevertheless, it can be seen that place and manner errors are mostly higher than voicing errors, although the total number of errors are higher than for human listeners. The results for the DNN in Fig. 4 show error patterns that are more similar to human perception results, with the most place errors and the least voicing errors. However, the overall error rates are higher and increase starting at higher SNRs.

\begin{figure}[htb]

\begin{minipage}[b]{.25\linewidth}
  \centering
  \centerline{\includegraphics[width=3.5cm]{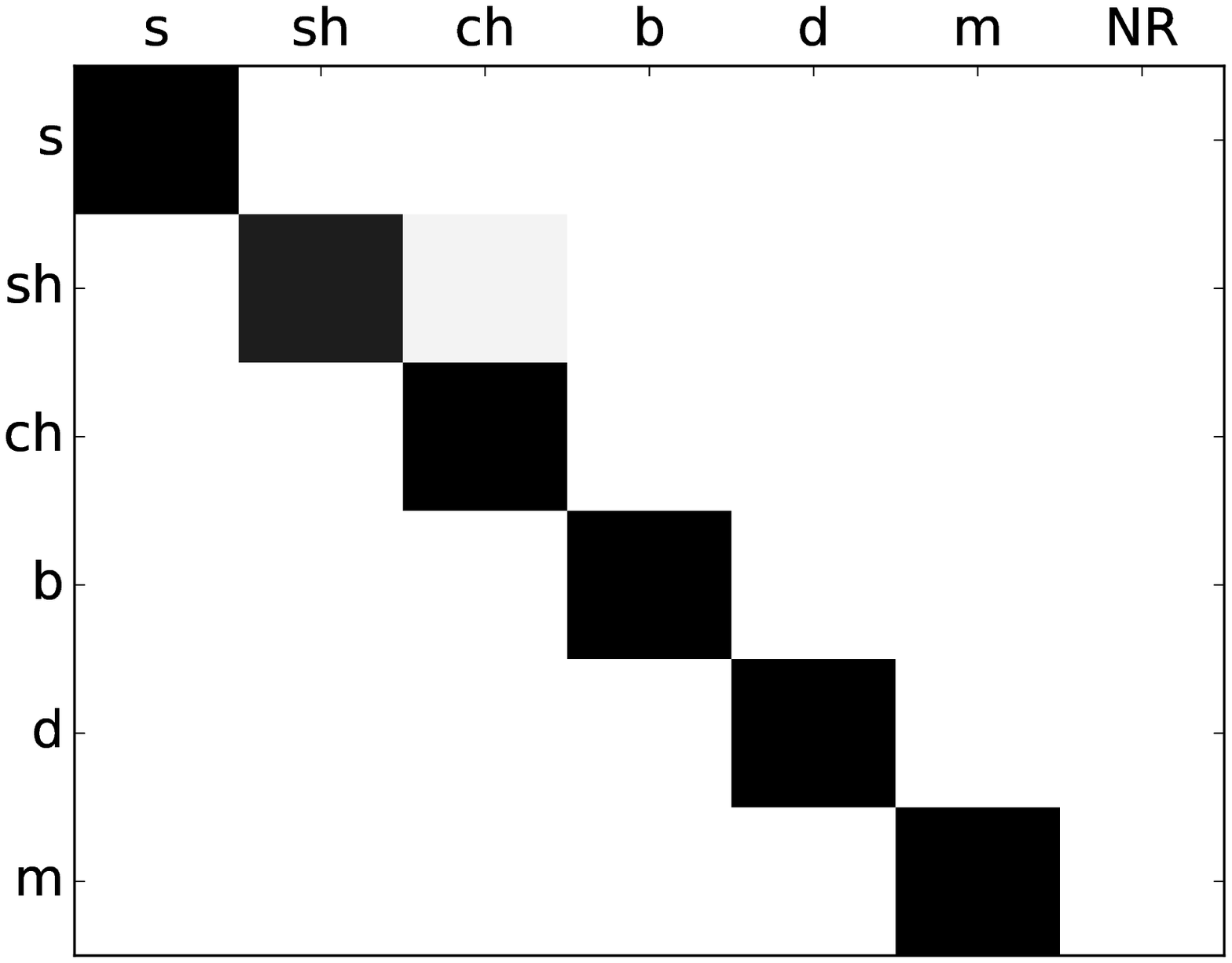}}
  \centerline{(a) SNR = 30dB}\medskip
\end{minipage}
\hspace{\fill}
\begin{minipage}[b]{0.25\linewidth}
  \centering
  \centerline{\includegraphics[width=3.5cm]{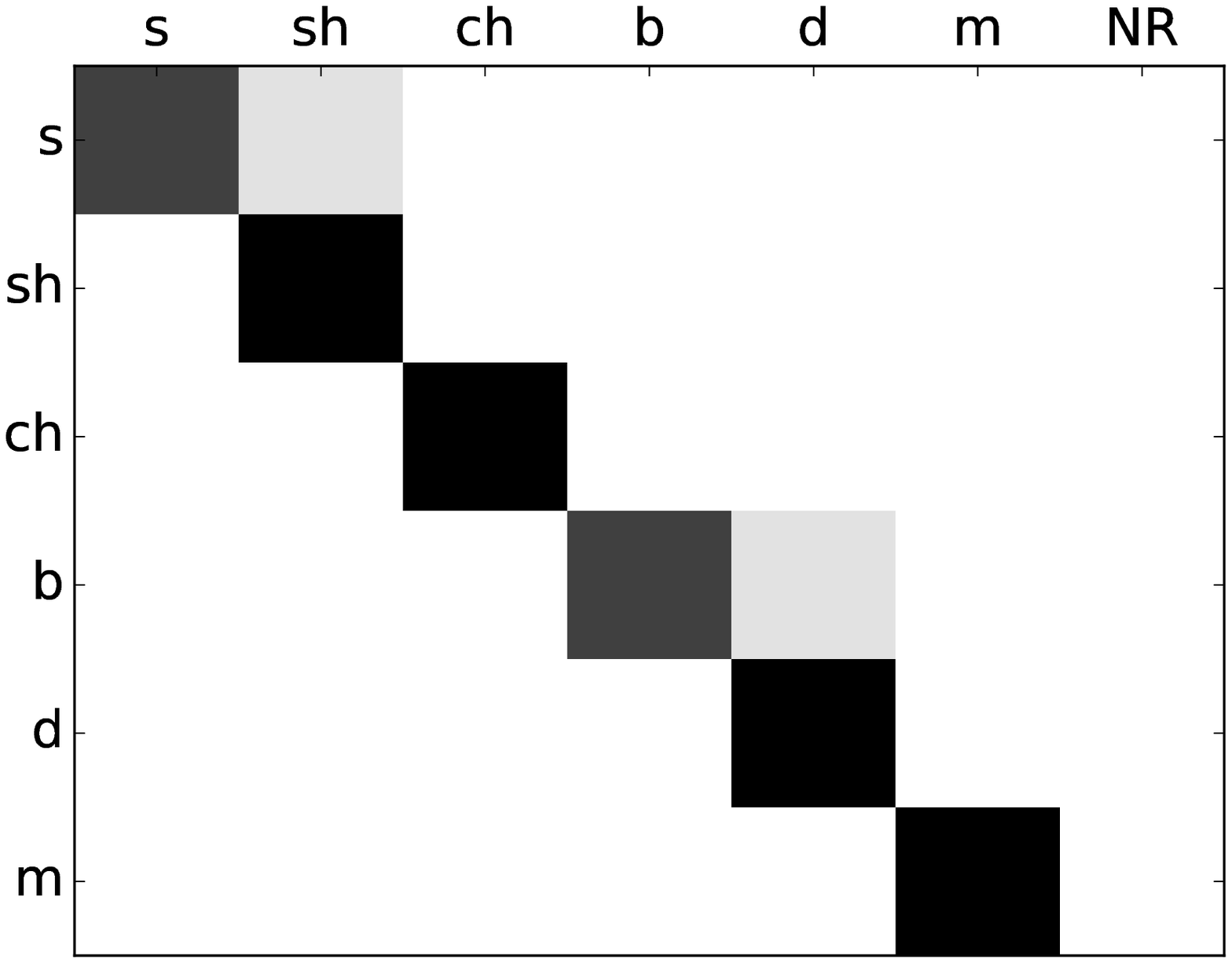}}
  \centerline{(b) SNR = 20dB}\medskip
\end{minipage}
\hfill
\begin{minipage}[b]{.25\linewidth}
  \centering
  \centerline{\includegraphics[width=3.5cm]{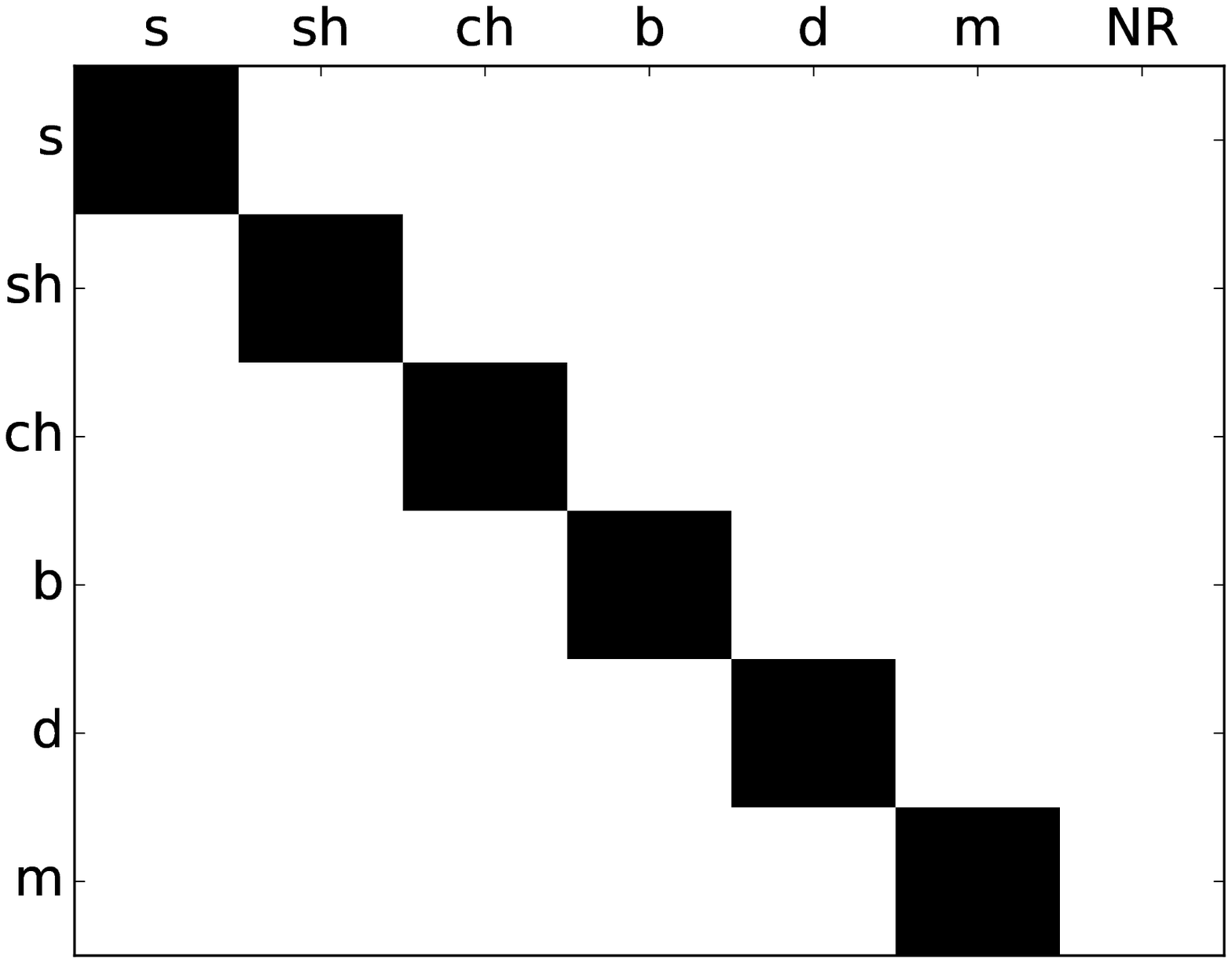}}
  \centerline{(c) SNR = 10dB}\medskip
\end{minipage}
\hfill
\begin{minipage}[b]{0.25\linewidth}
  \centering
  \centerline{\includegraphics[width=3cm]{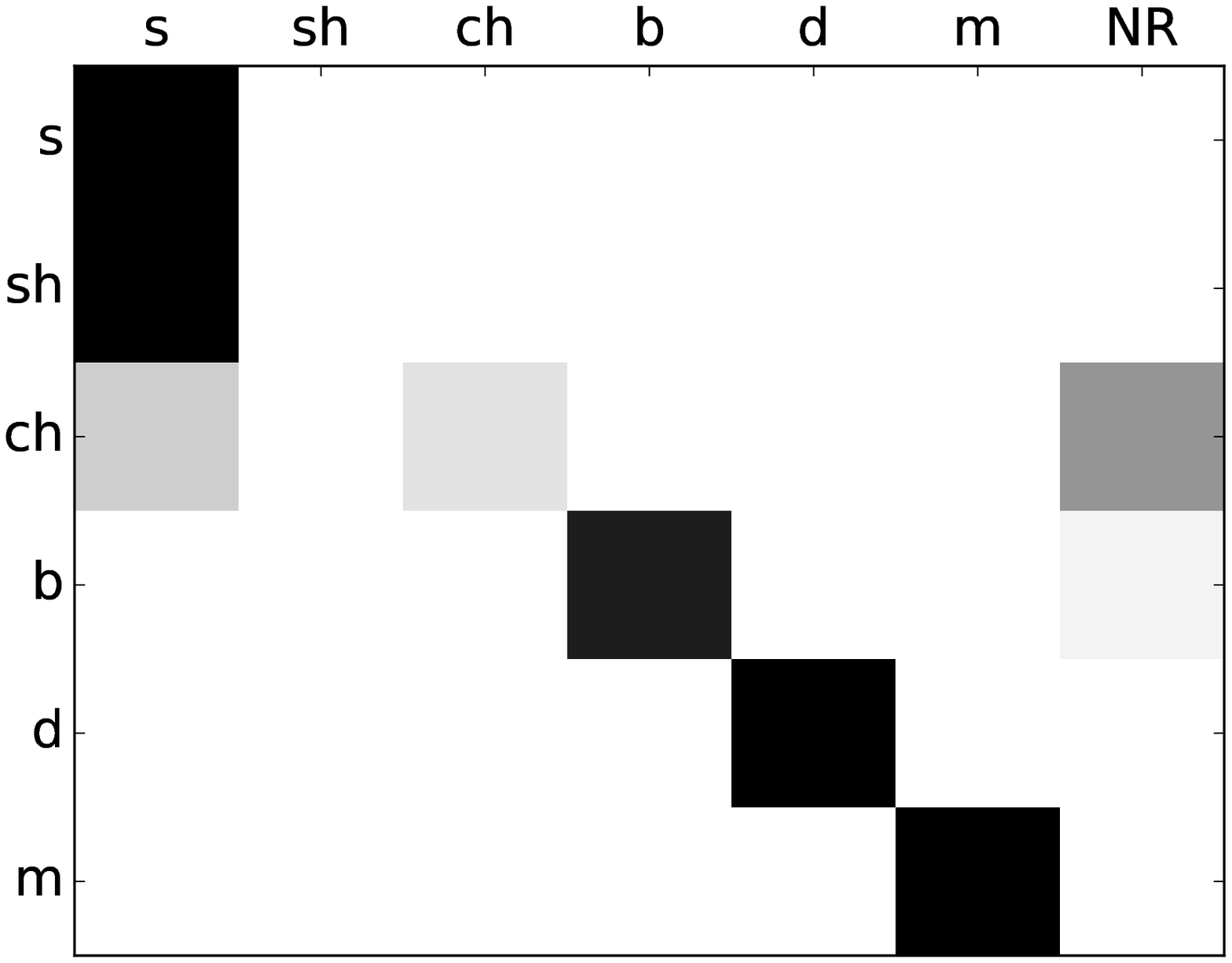}}
  \centerline{(d) SNR = 0dB}\medskip
\end{minipage}
\hfill
\begin{minipage}[b]{.25\linewidth}
  \centering
  \centerline{\includegraphics[width=3cm]{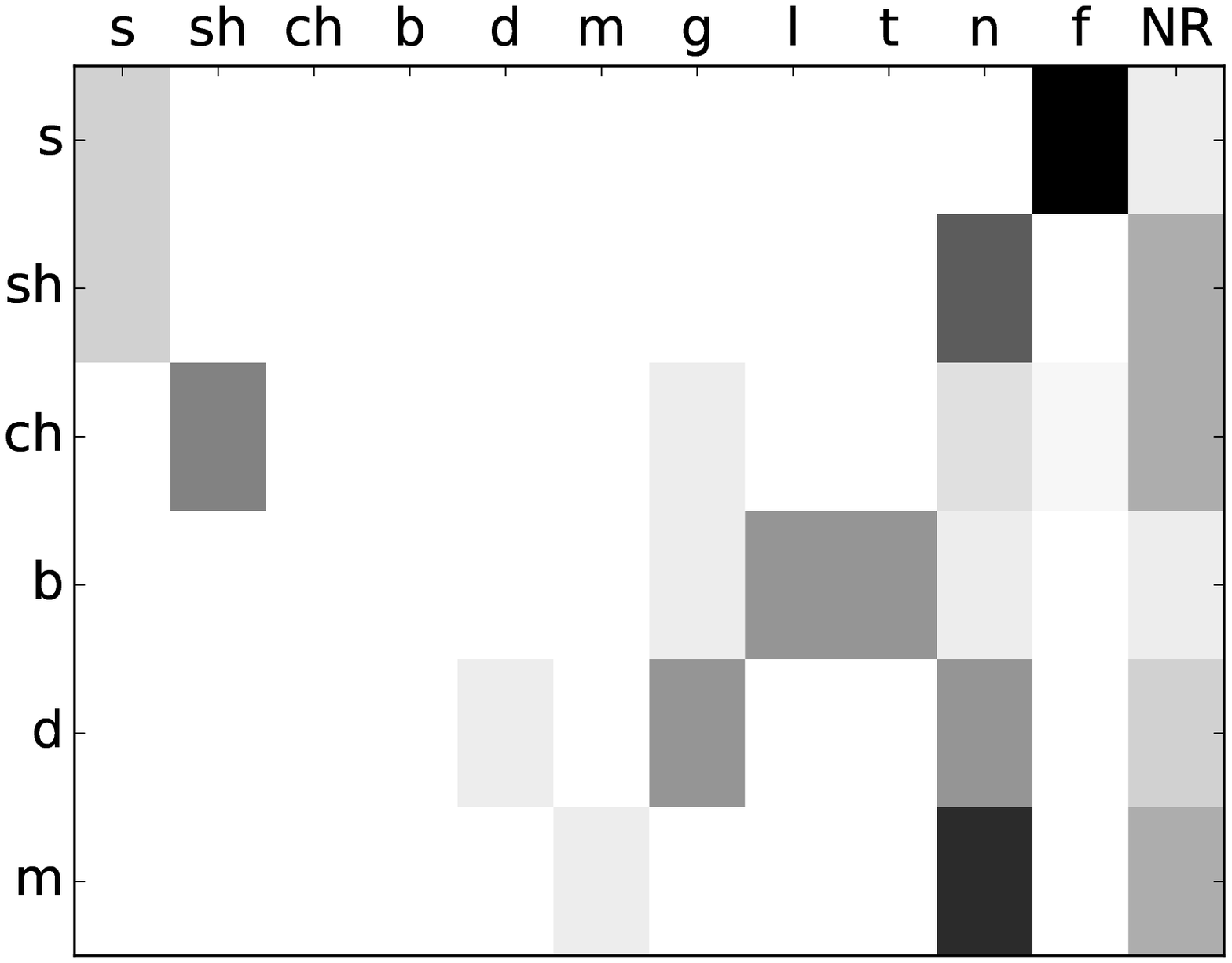}}
  \centerline{(e) SNR = -10dB}\medskip
\end{minipage}
\hfill
\begin{minipage}[b]{0.25\linewidth}
  \centering
  \centerline{\includegraphics[width=3cm]{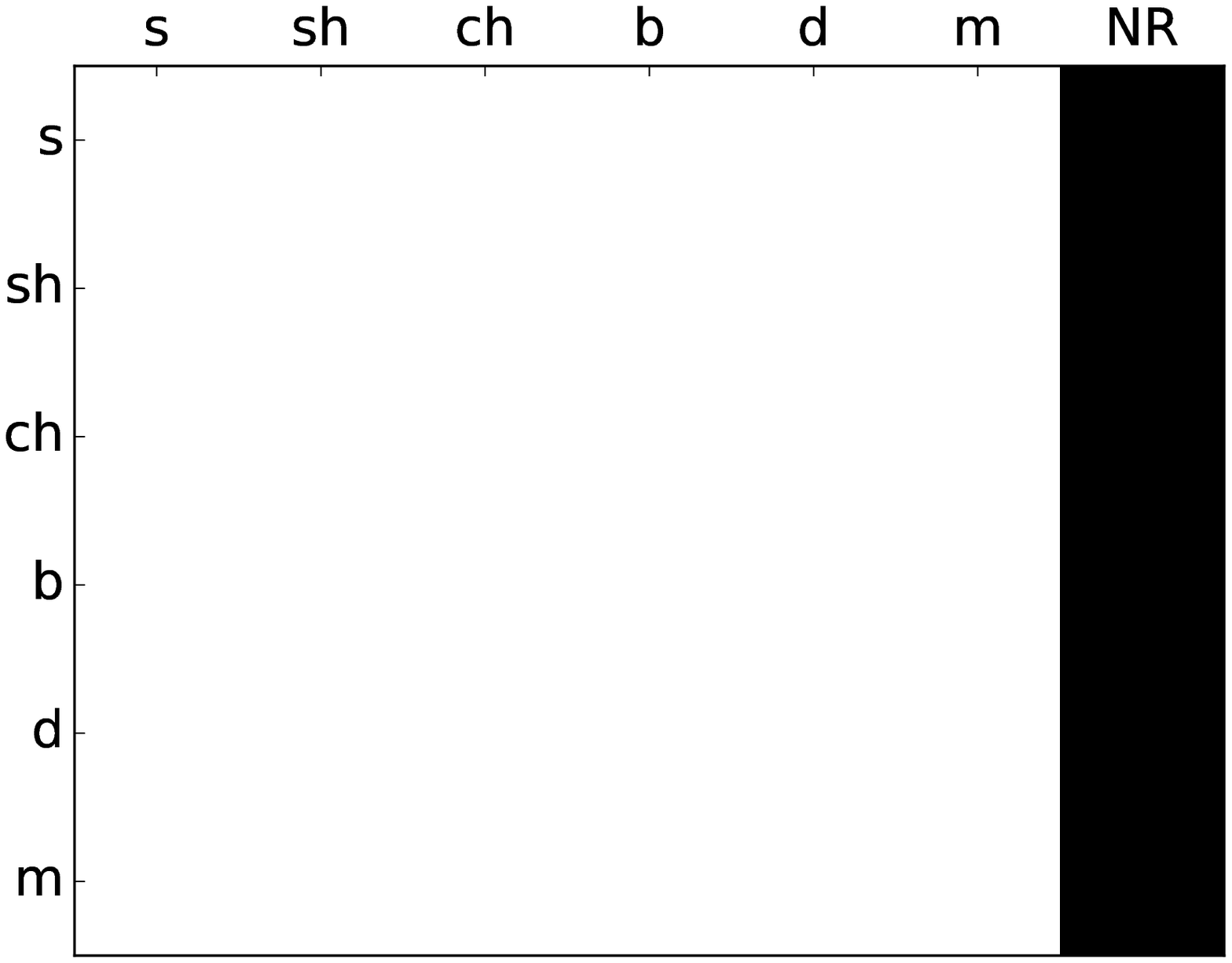}}
  \centerline{(f) SNR = -20dB}\medskip
\end{minipage}

\caption{Confusion matrices for human perception results}
\label{fig:cmhuman}
\end{figure}

\begin{figure}[htb]

\begin{minipage}[b]{.25\linewidth}
  \centering
  \centerline{\includegraphics[width=3.5cm]{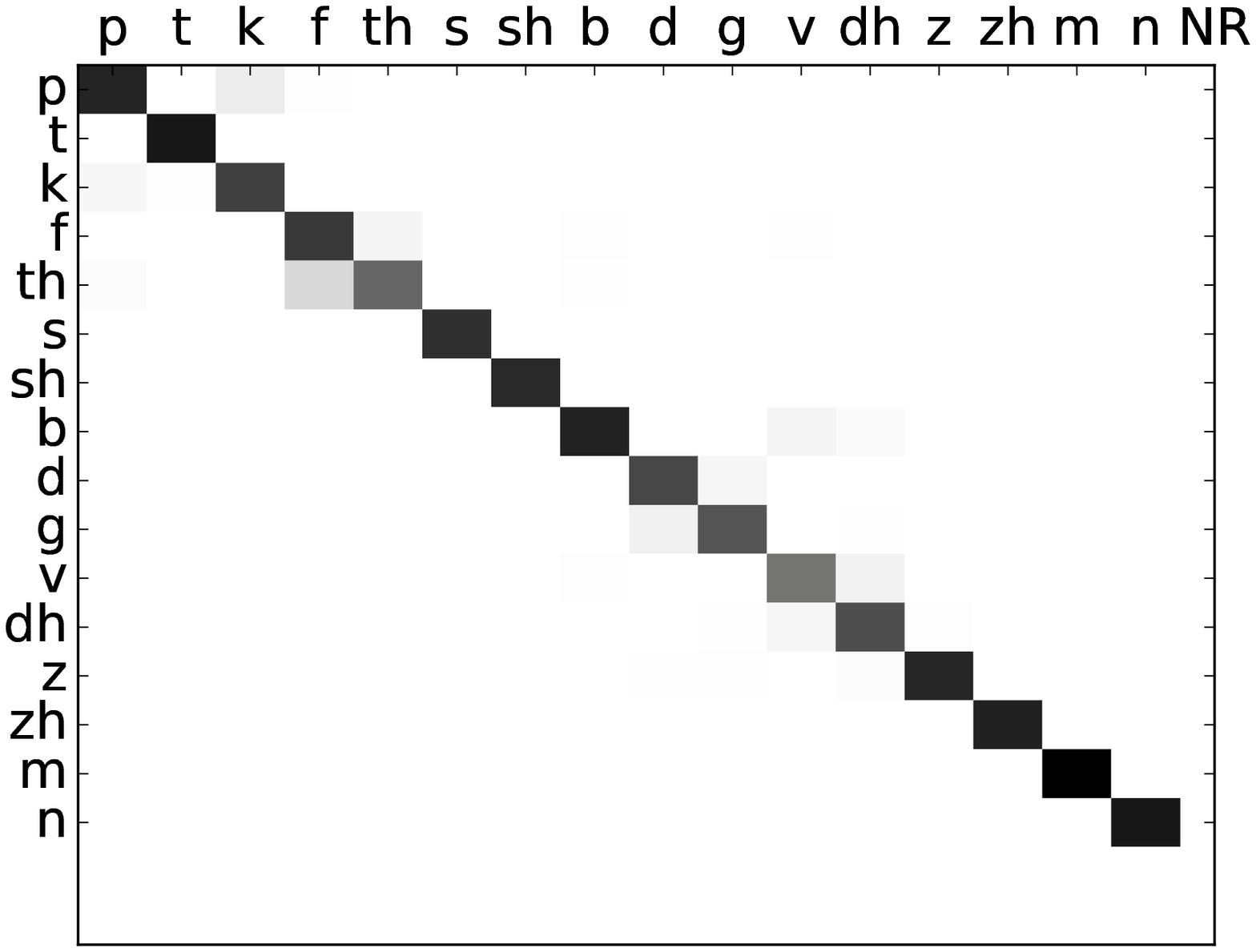}}
  \centerline{(a) SNR = 12dB}\medskip
\end{minipage}
\hspace{\fill}
\begin{minipage}[b]{0.25\linewidth}
  \centering
  \centerline{\includegraphics[width=3.4cm]{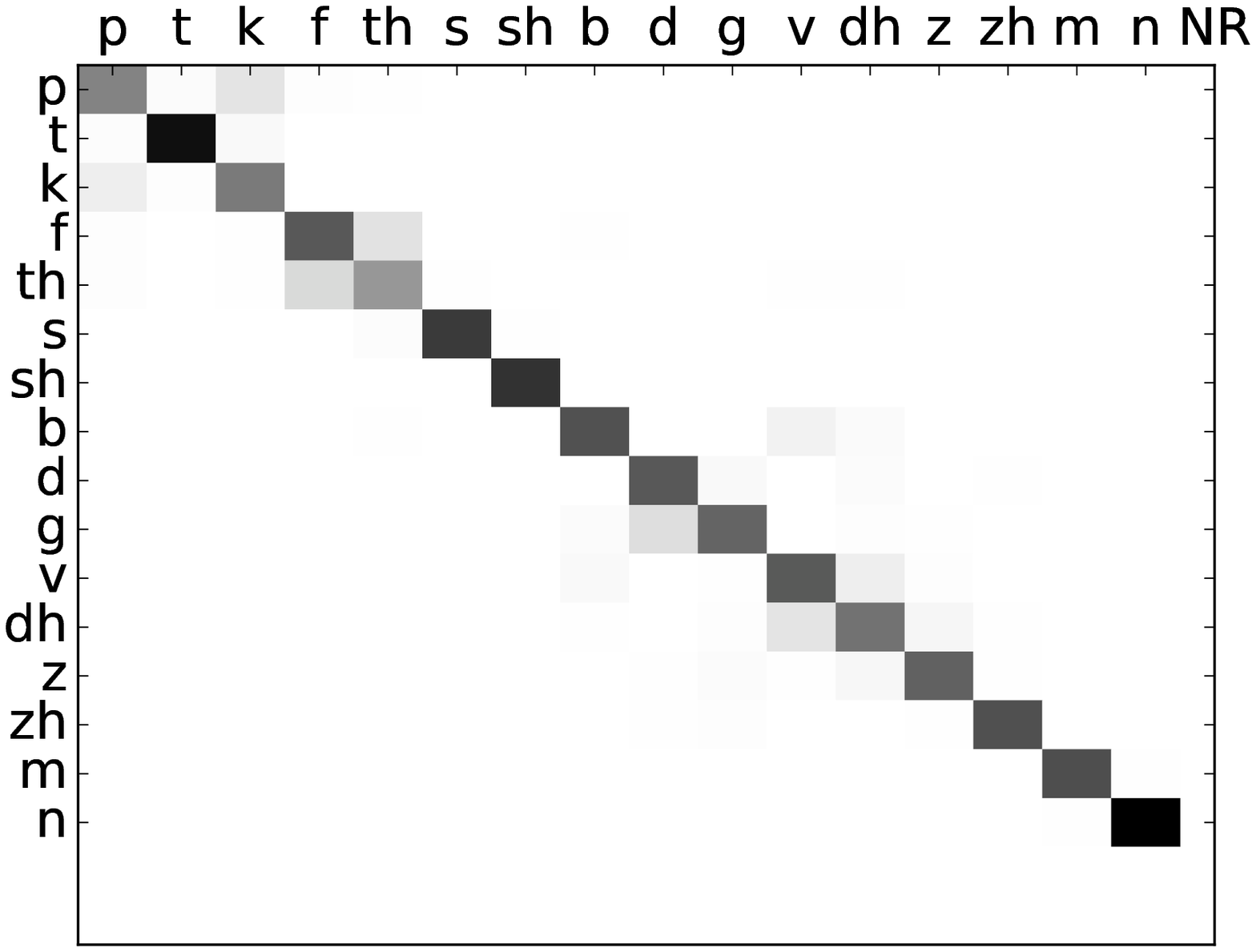}}
  \centerline{(b) SNR = 6dB}\medskip
\end{minipage}
\hfill
\begin{minipage}[b]{.25\linewidth}
  \centering
  \centerline{\includegraphics[width=3.5cm]{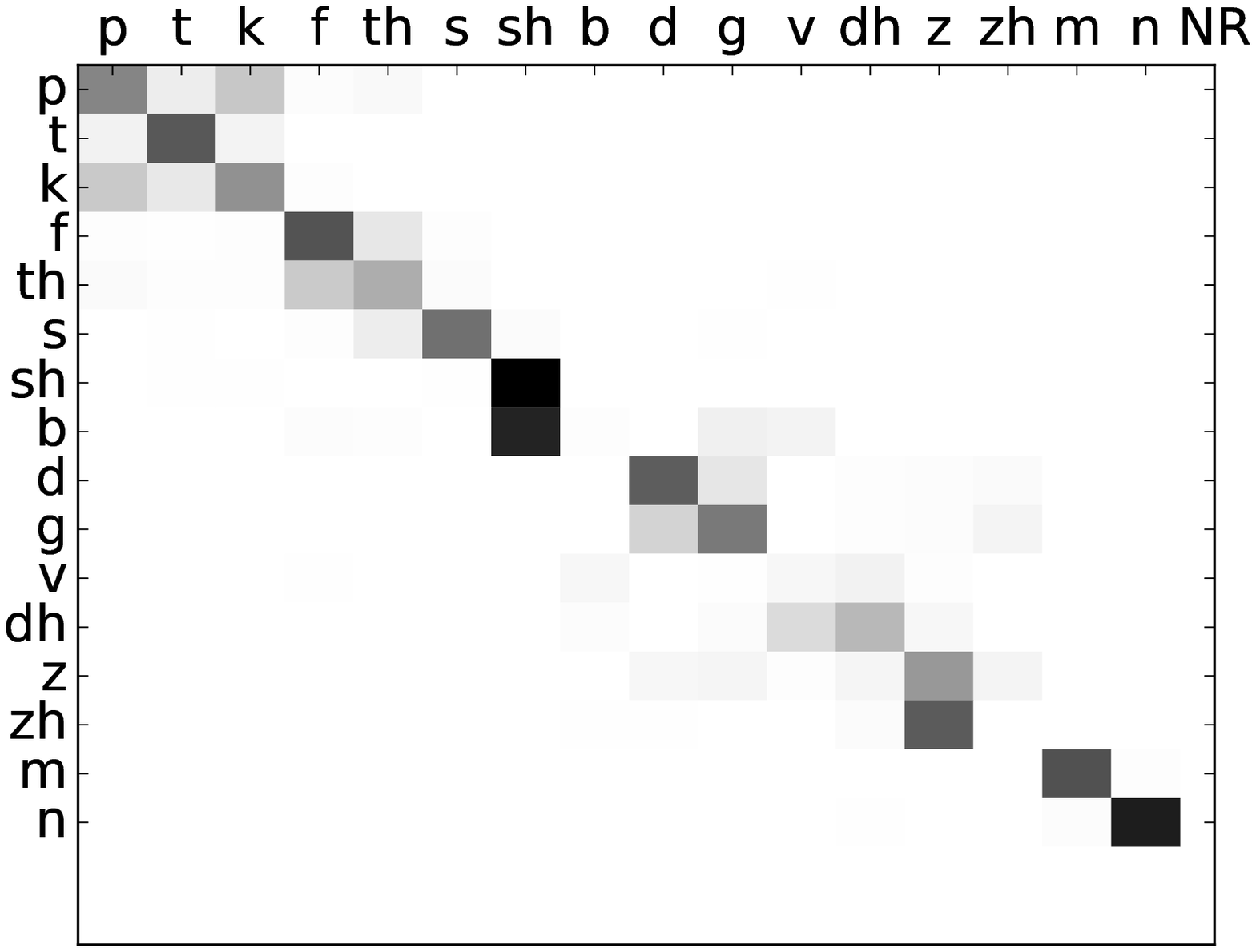}}
  \centerline{(c) SNR = 0dB}\medskip
\end{minipage}
\hfill
\begin{minipage}[b]{0.25\linewidth}
  \centering
  \centerline{\includegraphics[width=3.5cm]{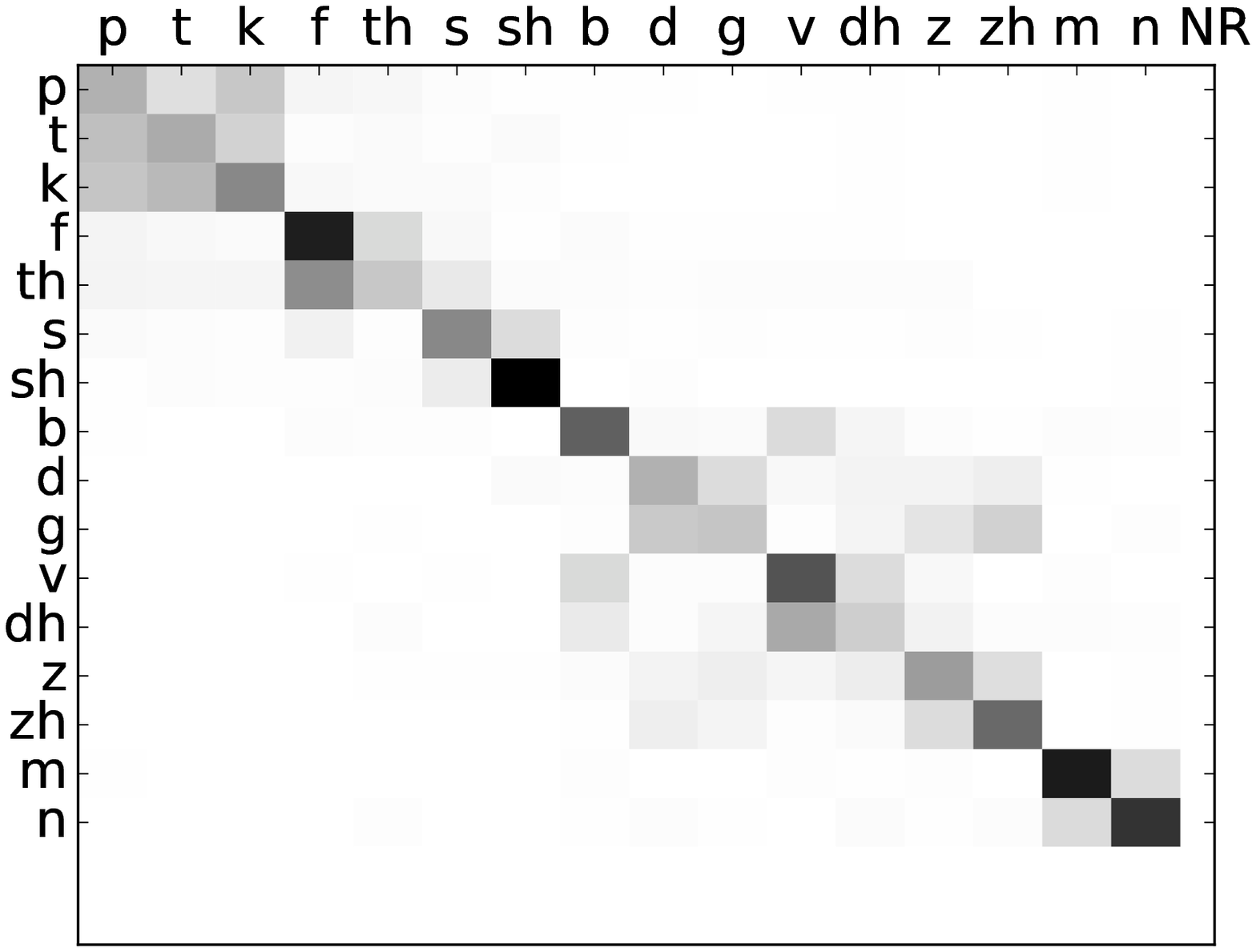}}
  \centerline{(d) SNR = -6dB}\medskip
\end{minipage}
\hfill
\begin{minipage}[b]{.25\linewidth}
  \centering
  \centerline{\includegraphics[width=3.5cm]{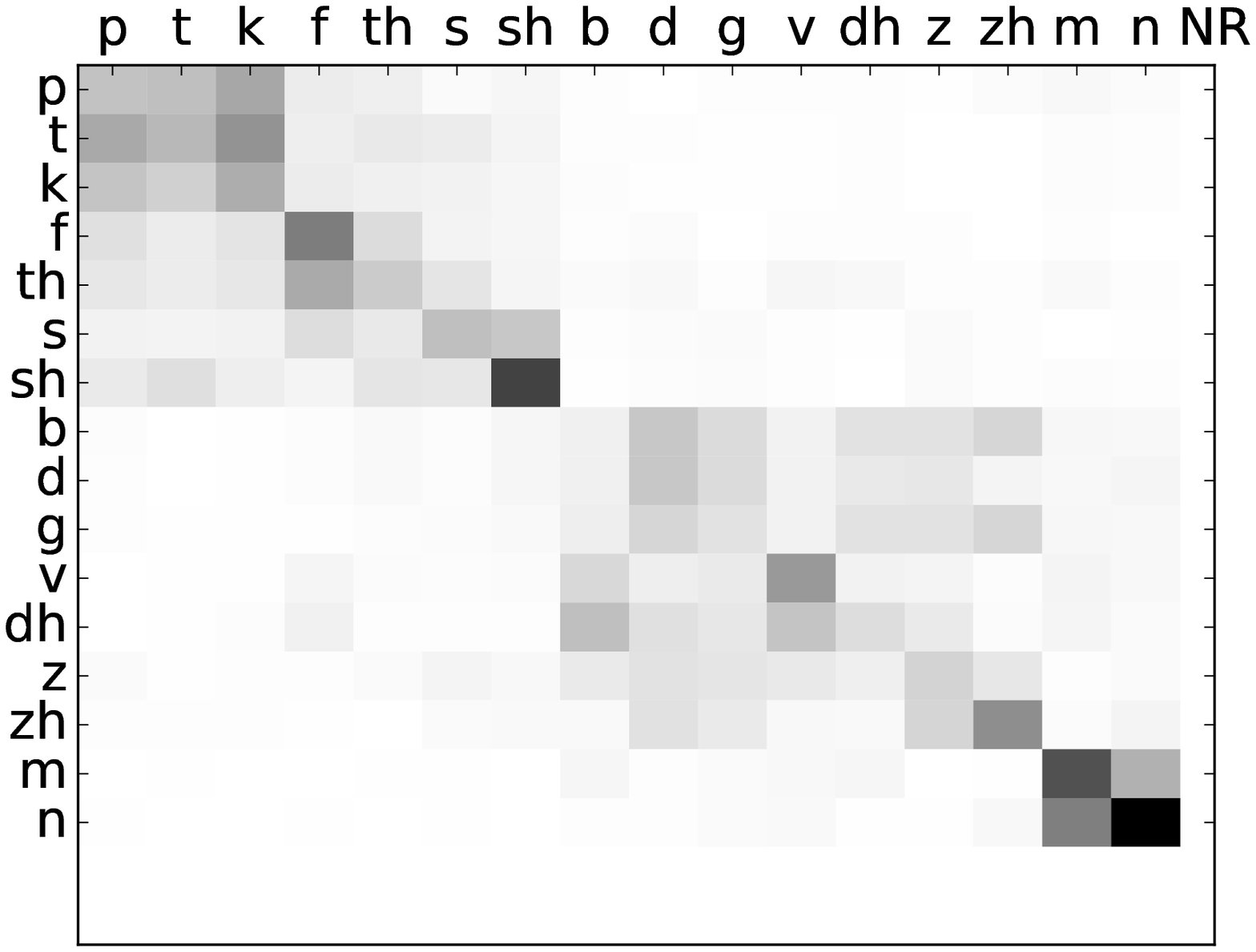}}
  \centerline{(e) SNR = -12dB}\medskip
\end{minipage}
\hfill
\begin{minipage}[b]{0.25\linewidth}
  \centering
  \centerline{\includegraphics[width=3.5cm]{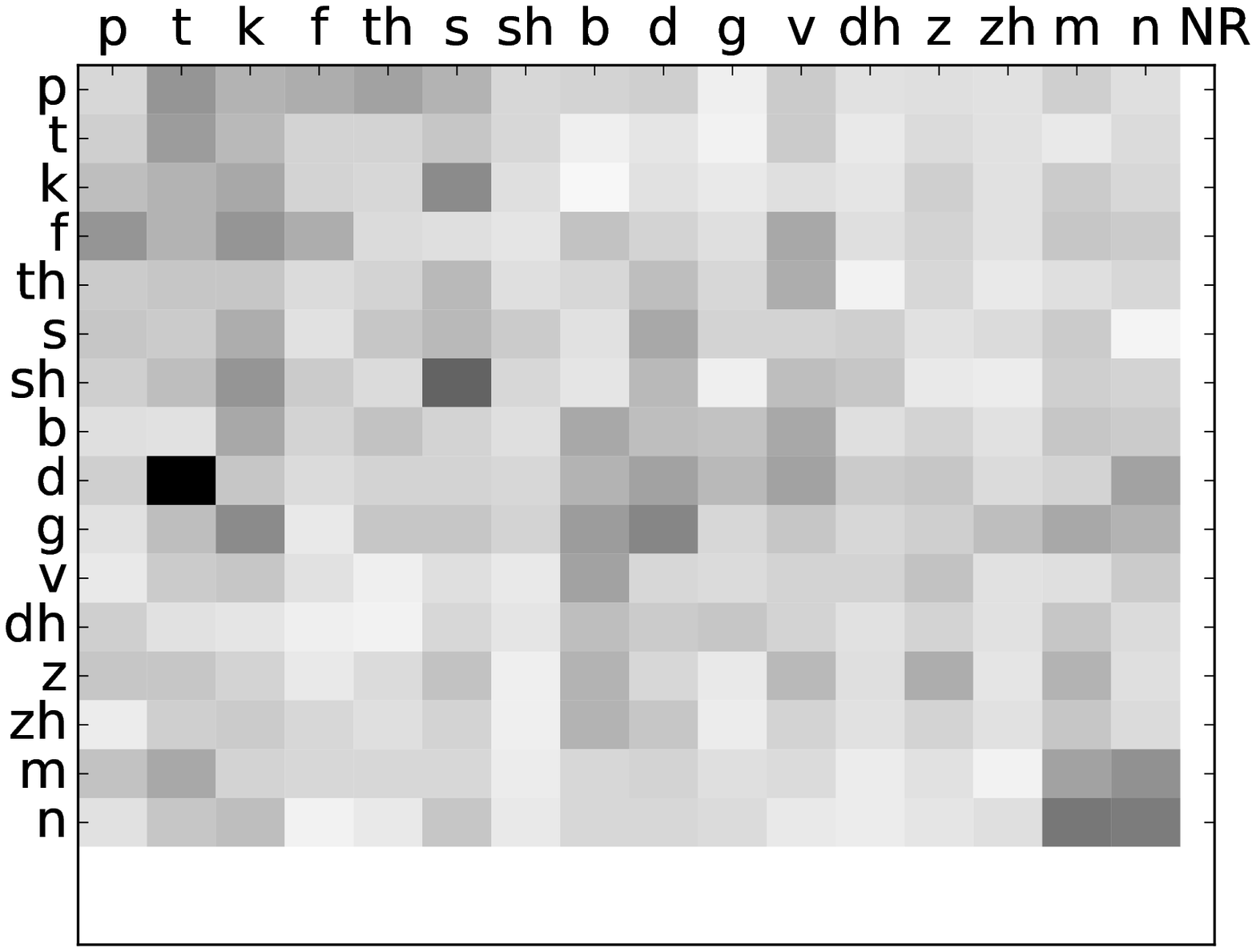}}
  \centerline{(f) SNR = -18dB}\medskip
\end{minipage}

\caption{Confusion matrices for Miller \& Nicely results}
\label{fig:cmmiller}
\end{figure}

\begin{figure*}[htb]

\begin{minipage}[b]{.13\linewidth}
  \centering
  \centerline{\includegraphics[width=2.5cm]{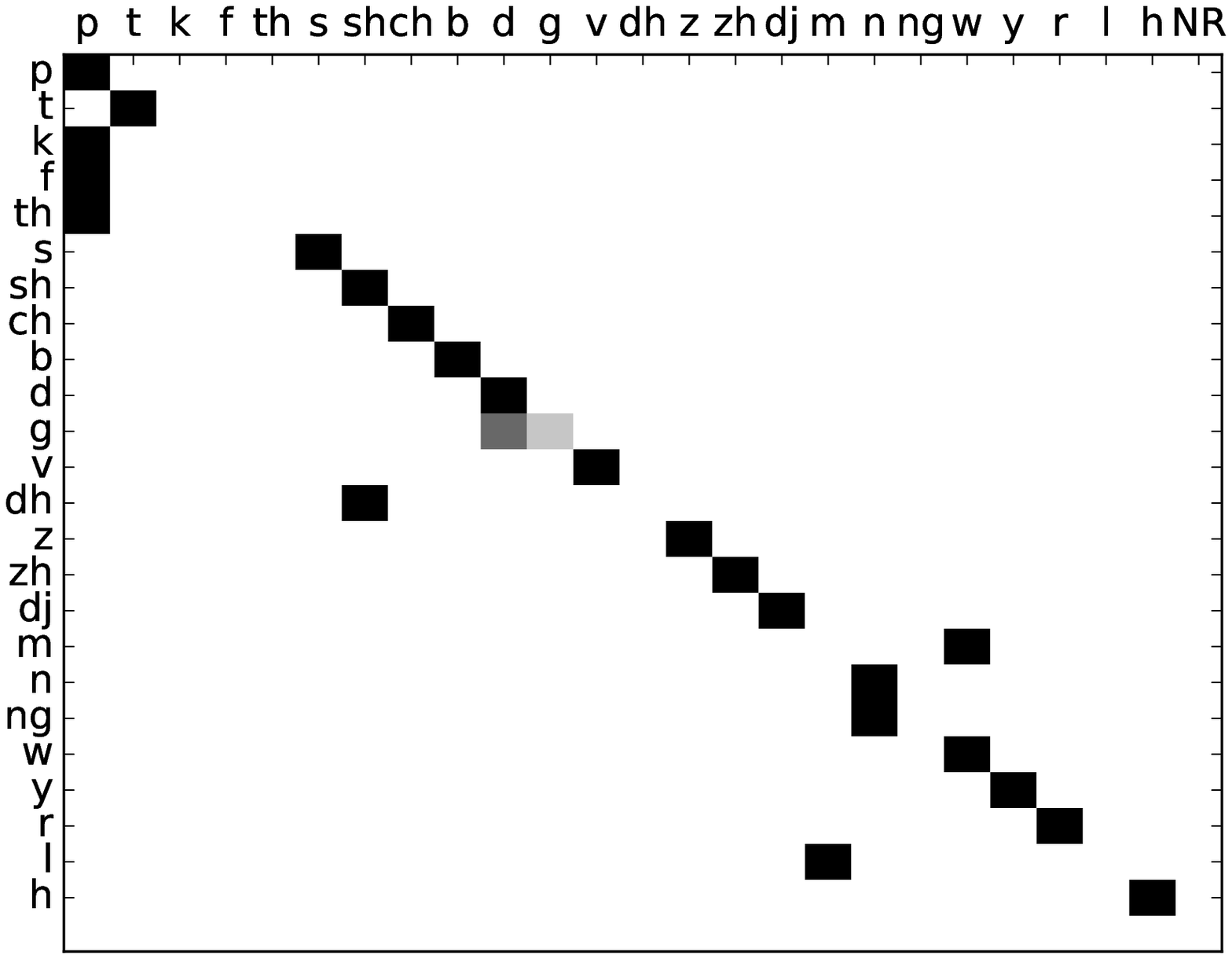}}
  \centerline{(a) SNR = 40dB}\medskip
\end{minipage}
\hspace{0.01cm}
\begin{minipage}[b]{0.13\linewidth}
  \centering
  \centerline{\includegraphics[width=2.5cm]{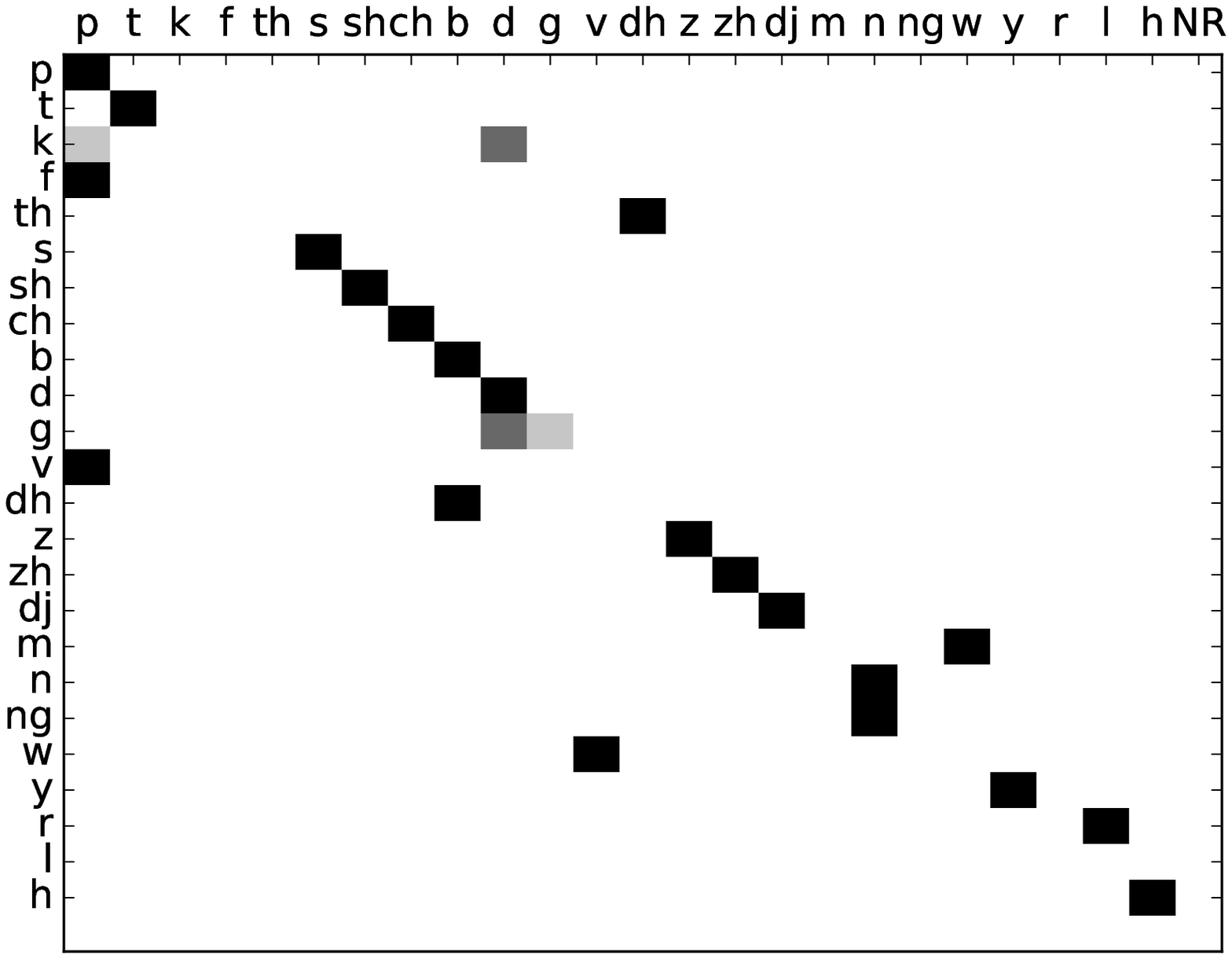}}
  \centerline{(b) SNR = 30dB}\medskip
\end{minipage}
\hspace{0.01cm}
\begin{minipage}[b]{.13\linewidth}
  \centering
  \centerline{\includegraphics[width=2.5cm]{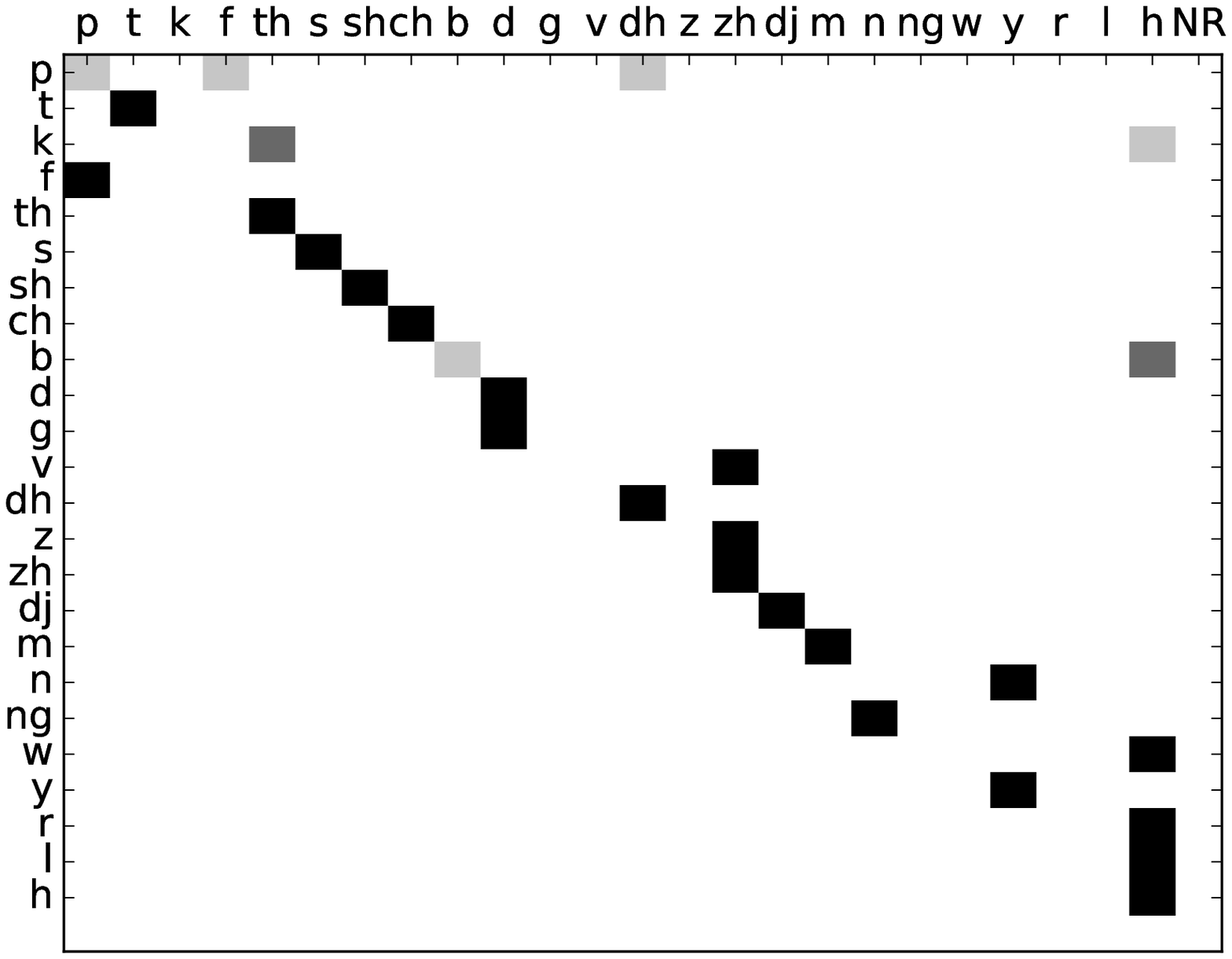}}
  \centerline{(c) SNR = 20dB}\medskip
\end{minipage}
\hspace{0.01cm}
\begin{minipage}[b]{0.13\linewidth}
  \centering
  \centerline{\includegraphics[width=2.5cm]{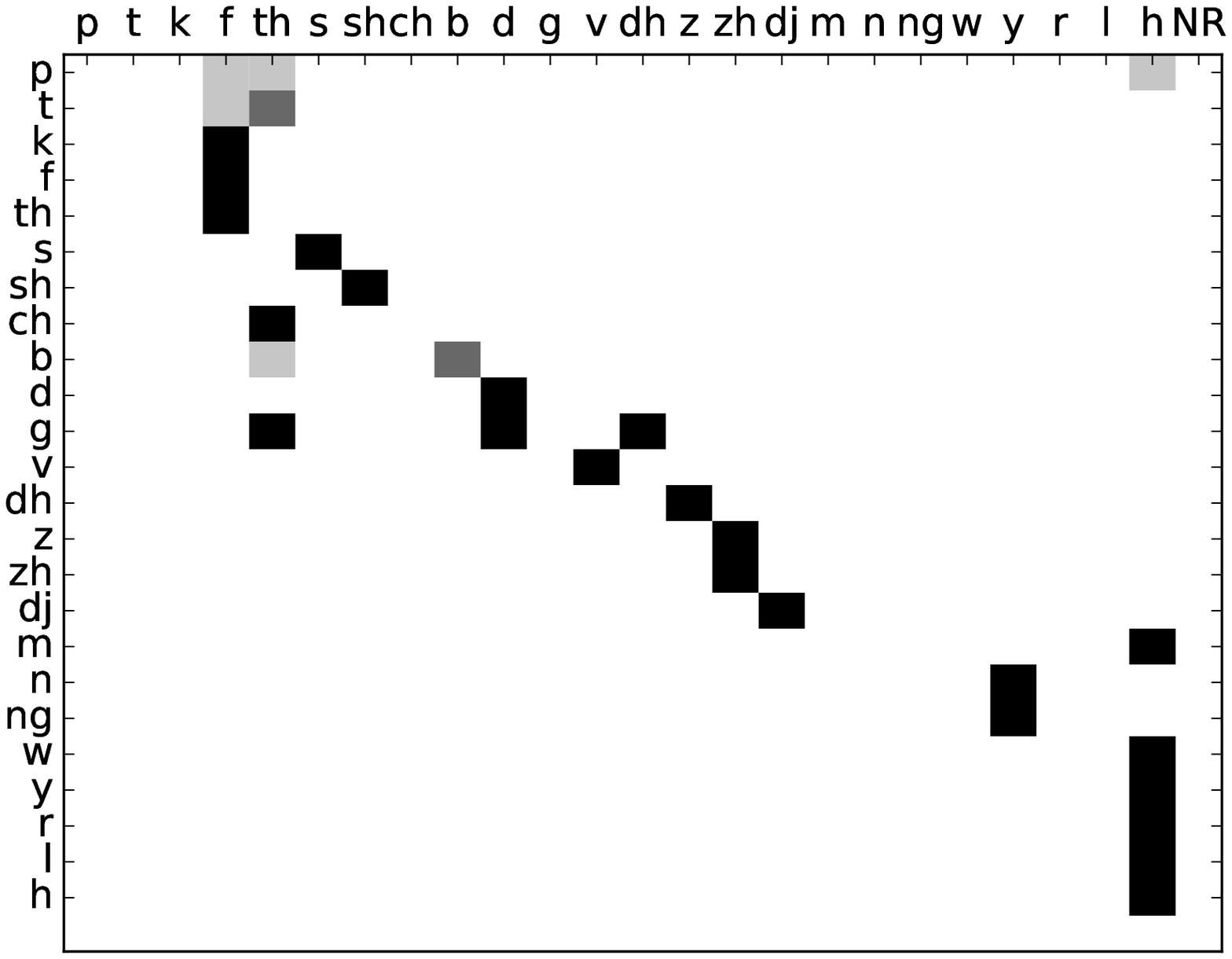}}
  \centerline{(d) SNR = 10dB}\medskip
\end{minipage}
\hspace{0.01cm}
\begin{minipage}[b]{.13\linewidth}
  \centering
  \centerline{\includegraphics[width=2.5cm]{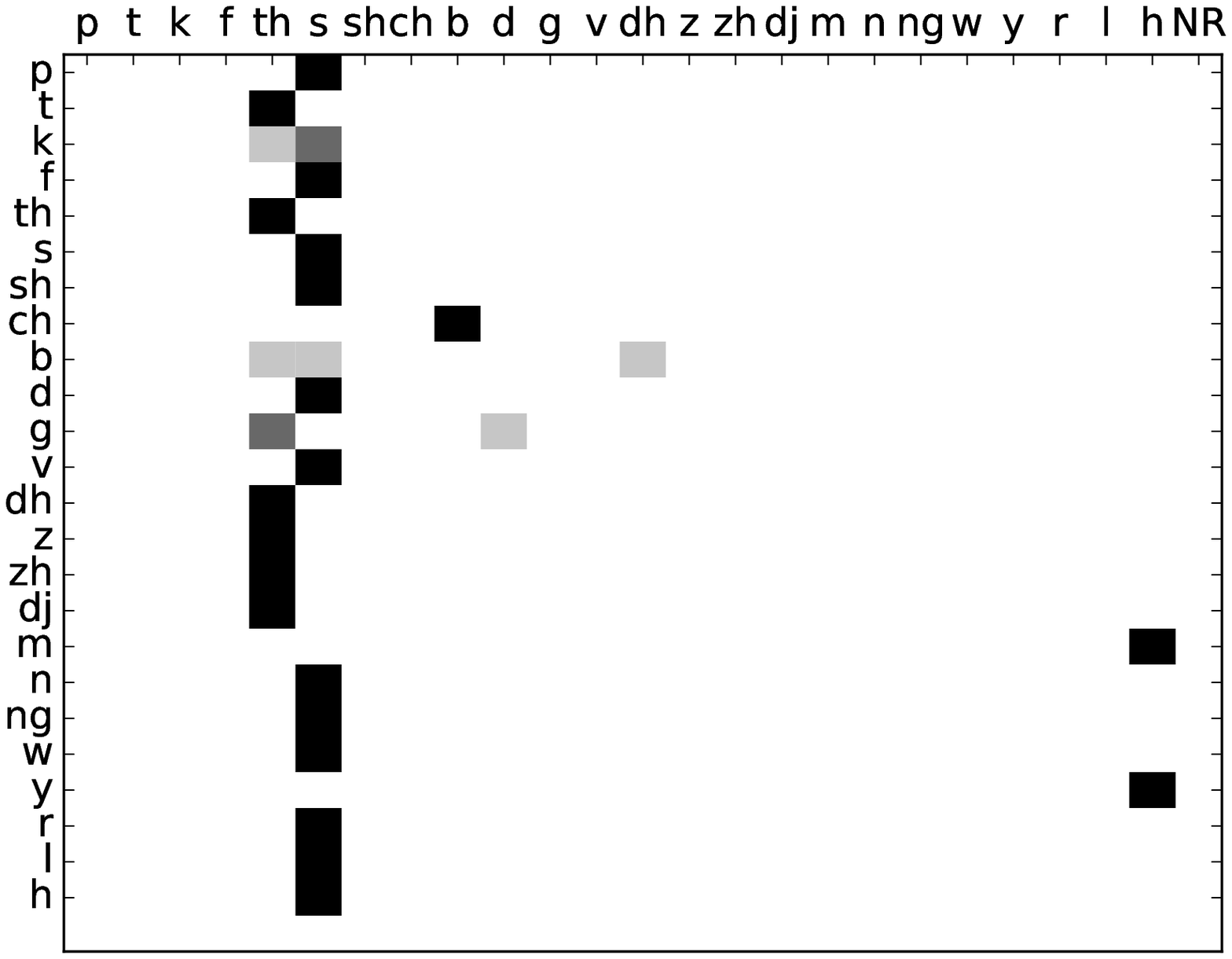}}
  \centerline{(e) SNR = 0dB}\medskip
\end{minipage}
\hspace{0.01cm}
\begin{minipage}[b]{0.13\linewidth}
  \centering
  \centerline{\includegraphics[width=2.5cm]{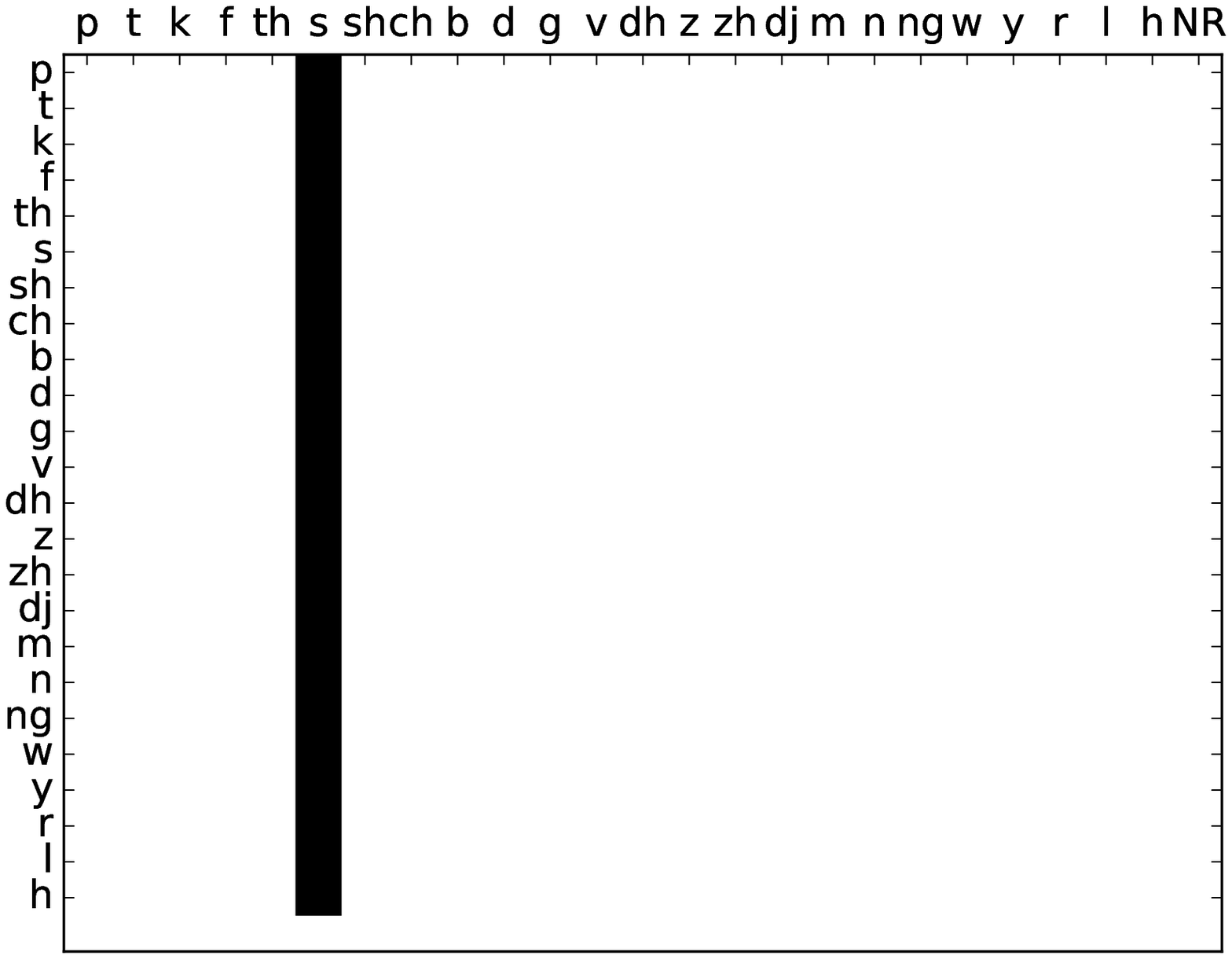}}
  \centerline{(f) SNR = -10dB}\medskip
\end{minipage}
\hspace{0.01cm}
\begin{minipage}[b]{.13\linewidth}
  \centering
  \centerline{\includegraphics[width=2.5cm]{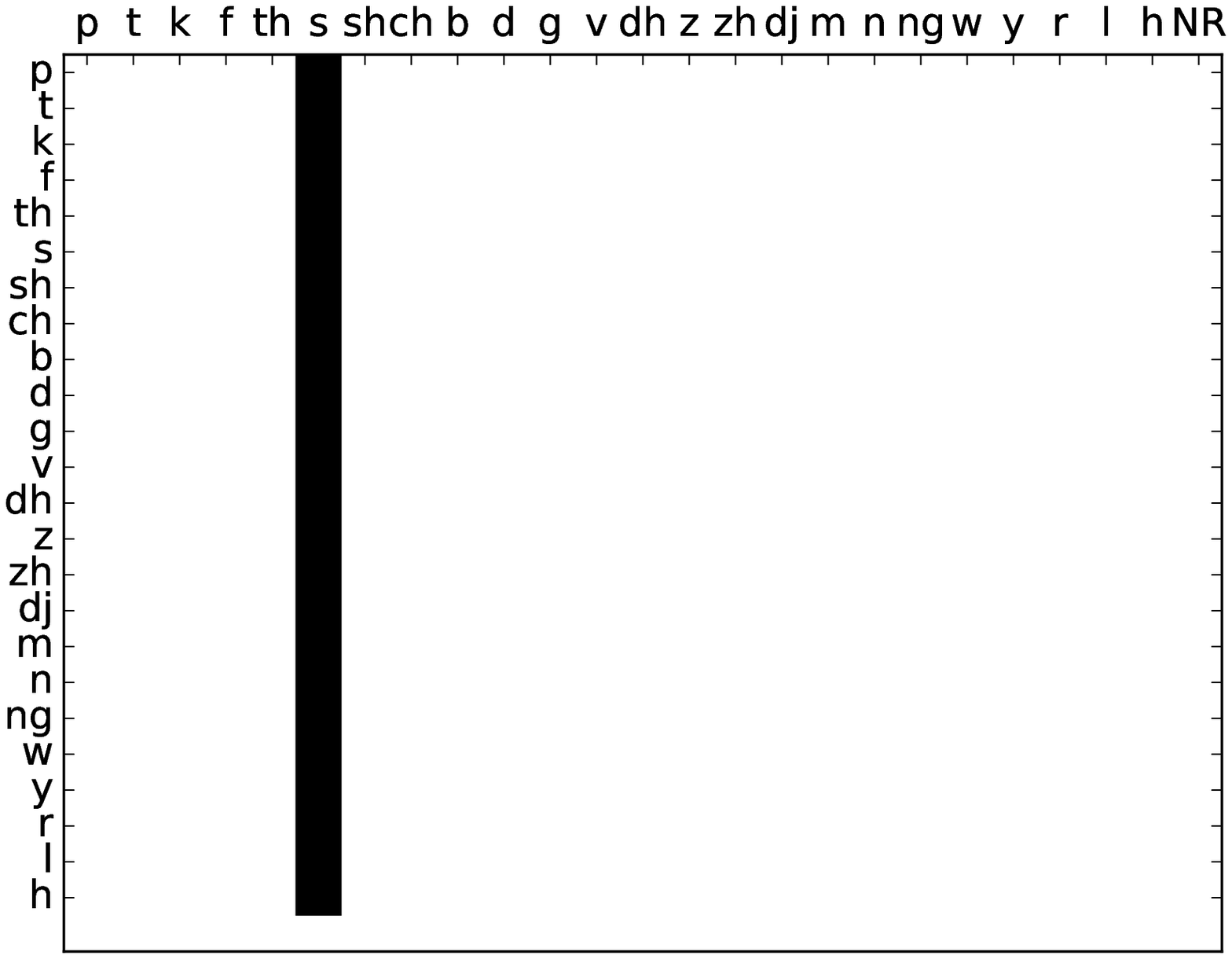}}
  \centerline{(g) SNR = -20dB}\medskip
\end{minipage}
\hfill
\caption{Confusion matrices for HMM-based system results}
\label{fig:cmhmm}
\end{figure*}

\begin{figure*}[htb]

\begin{minipage}[b]{.13\linewidth}
  \centering
  \centerline{\includegraphics[width=2.5cm]{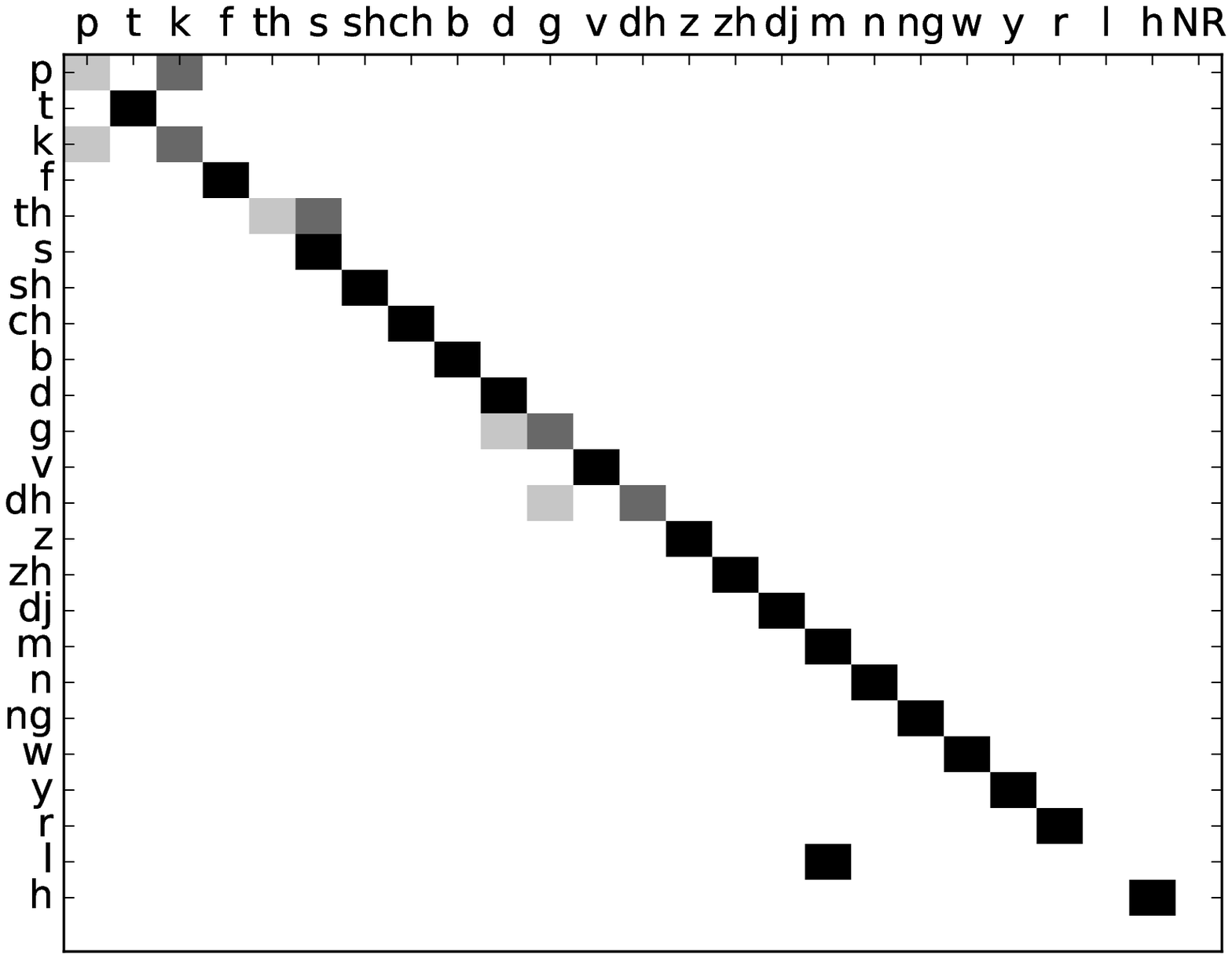}}
  \centerline{(a) SNR = 40dB}\medskip
\end{minipage}
\hspace{0.01cm}
\begin{minipage}[b]{.13\linewidth}
  \centering
  \centerline{\includegraphics[width=2.5cm]{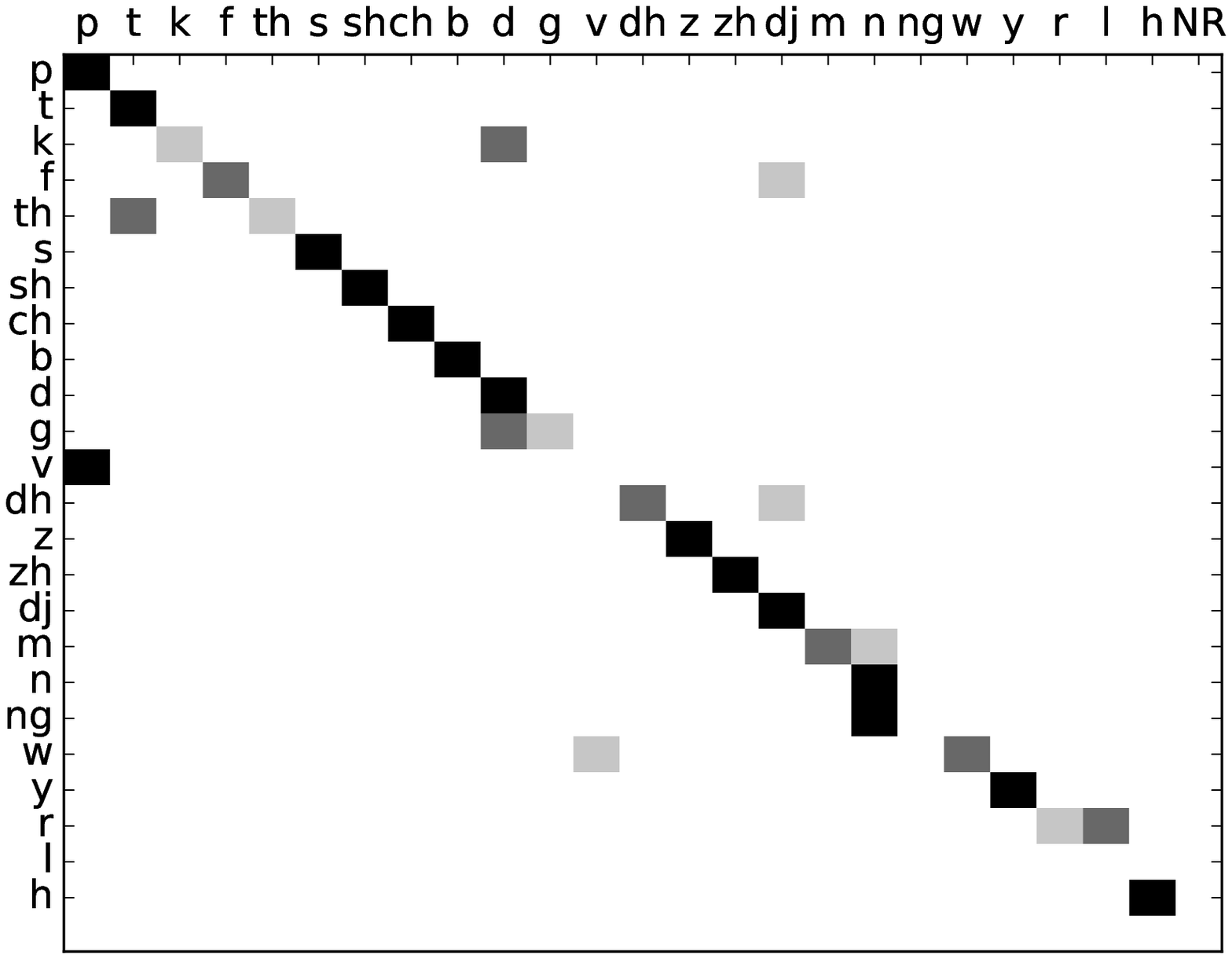}}
  \centerline{(b) SNR = 30dB}\medskip
\end{minipage}
\hspace{0.01cm}
\begin{minipage}[b]{.13\linewidth}
  \centering
  \centerline{\includegraphics[width=2.5cm]{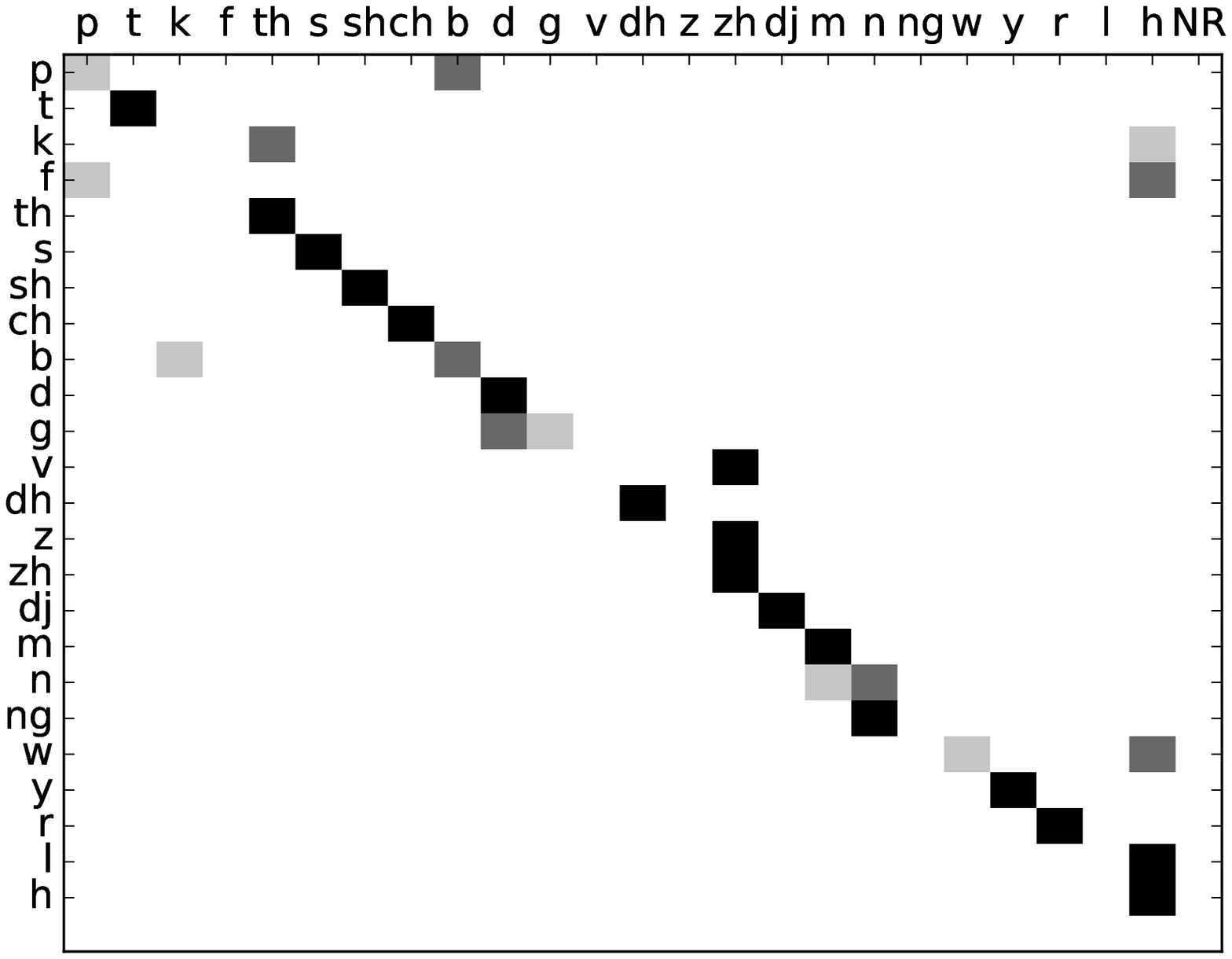}}
  \centerline{(c) SNR = 20dB}\medskip
\end{minipage}
\hspace{0.01cm}
\begin{minipage}[b]{.13\linewidth}
  \centering
  \centerline{\includegraphics[width=2.5cm]{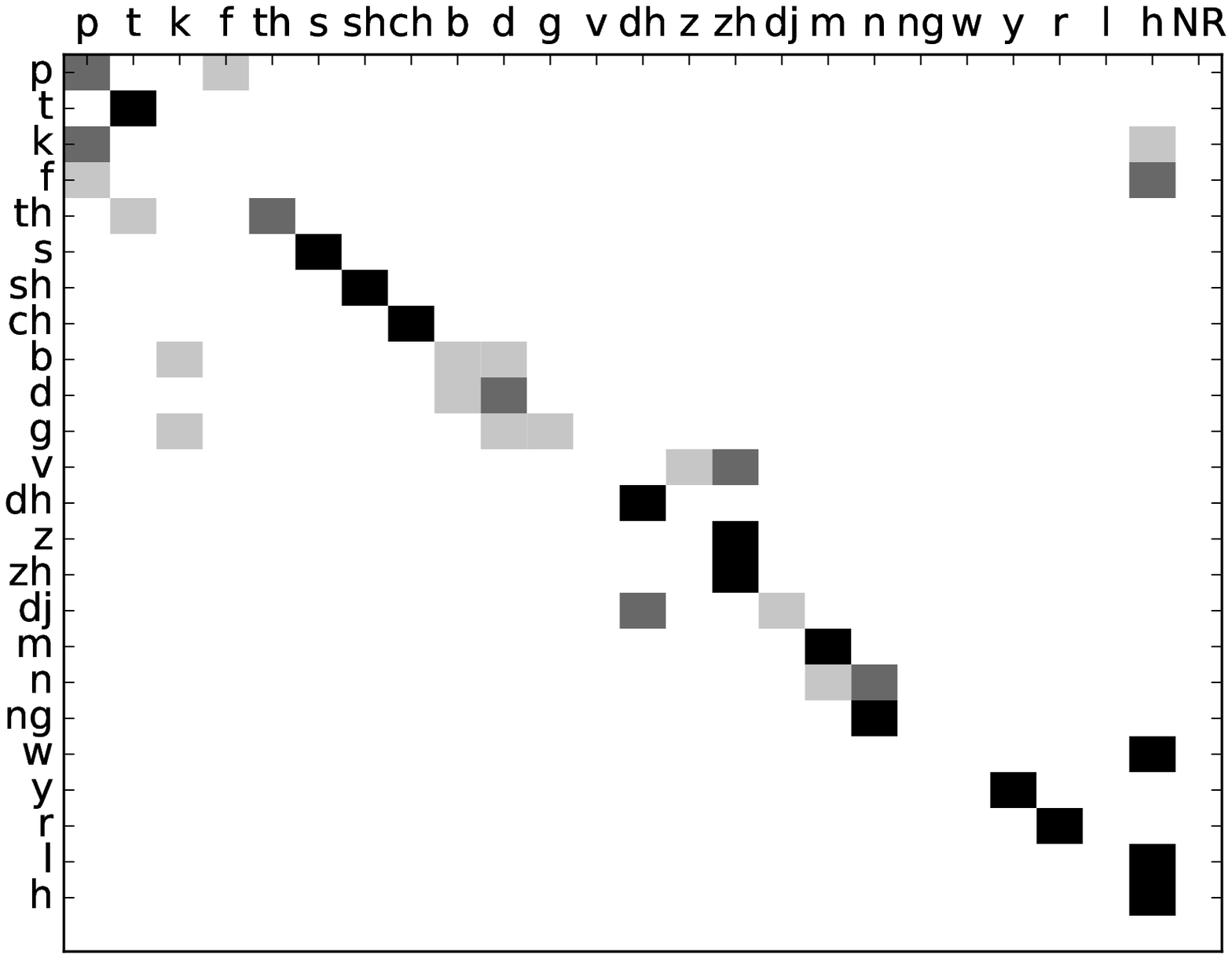}}
  \centerline{(d) SNR = 10dB}\medskip
\end{minipage}
\hspace{0.01cm}
\begin{minipage}[b]{.13\linewidth}
  \centering
  \centerline{\includegraphics[width=2.5cm]{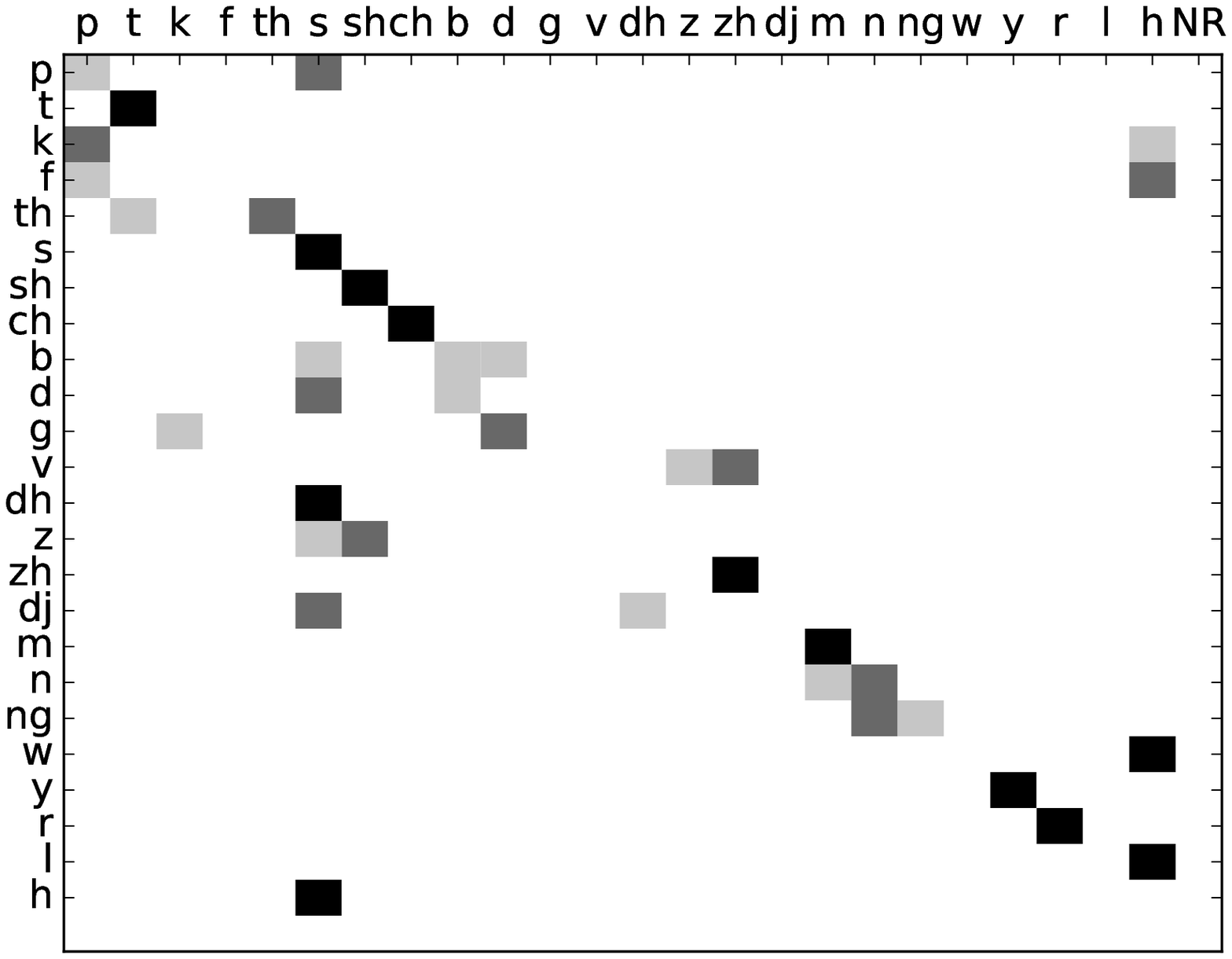}}
  \centerline{(e) SNR = 0dB}\medskip
\end{minipage}
\hspace{0.01cm}
\begin{minipage}[b]{.13\linewidth}
  \centering
  \centerline{\includegraphics[width=2.5cm]{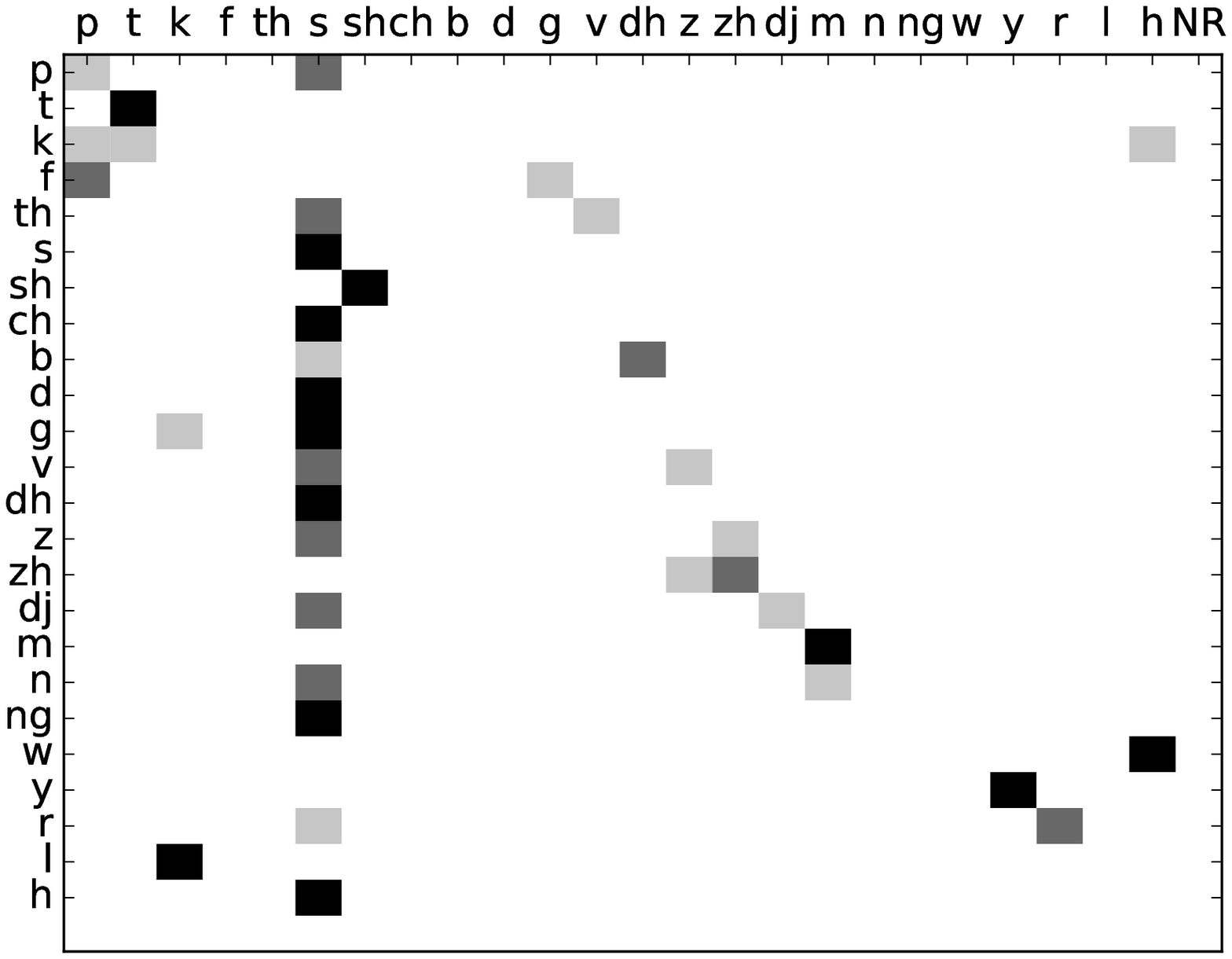}}
  \centerline{(f) SNR = -10dB}\medskip
\end{minipage}
\hspace{0.01cm}
\begin{minipage}[b]{.13\linewidth}
  \centering
  \centerline{\includegraphics[width=2.5cm]{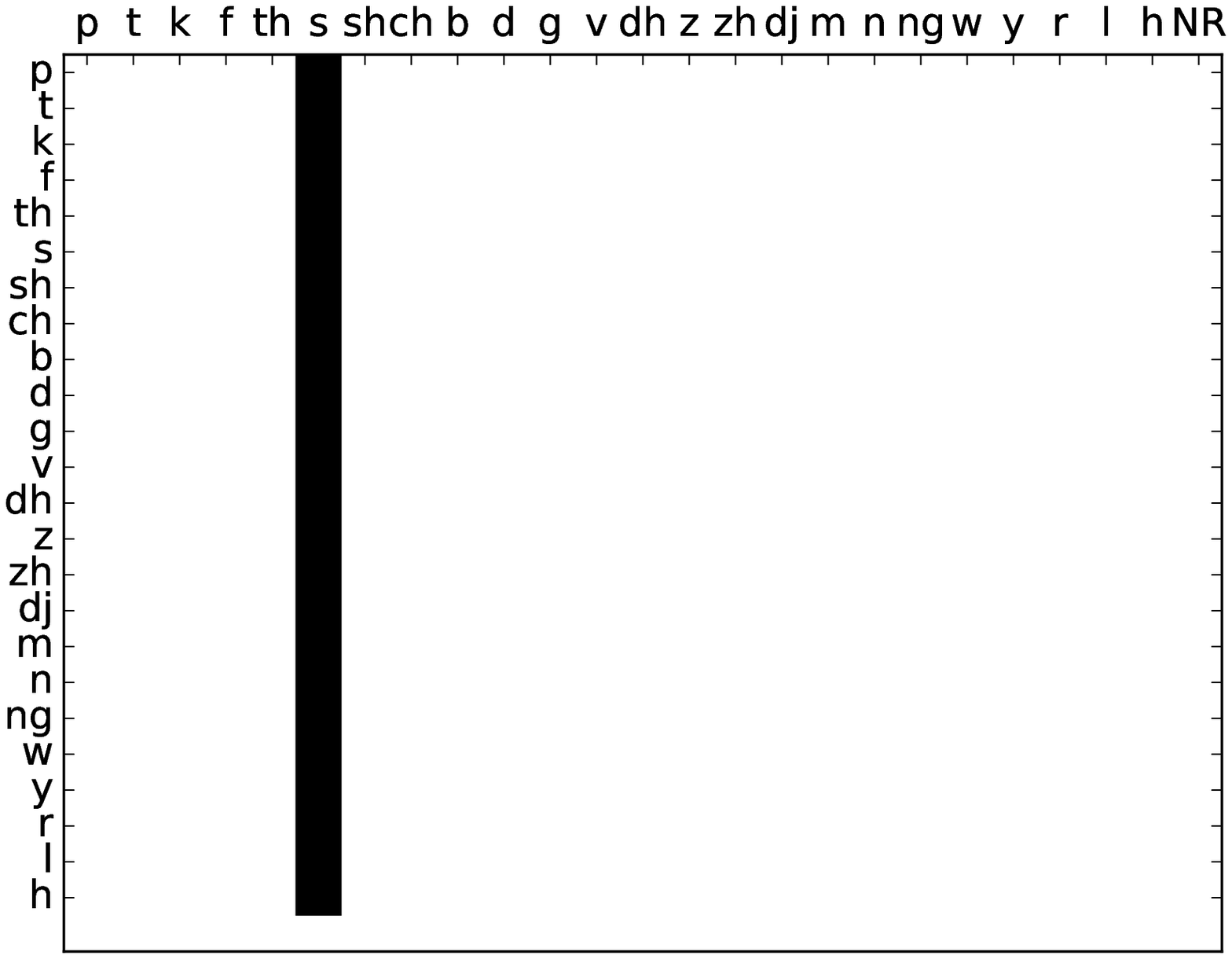}}
  \centerline{(g) SNR = -20dB}\medskip
\end{minipage}
\hfill
\caption{Confusion matrices for DNN-based system results}
\label{fig:cmdnn}
\end{figure*}

\begin{figure}[!htb]
\centering
\includegraphics[width=8cm,height=4cm]{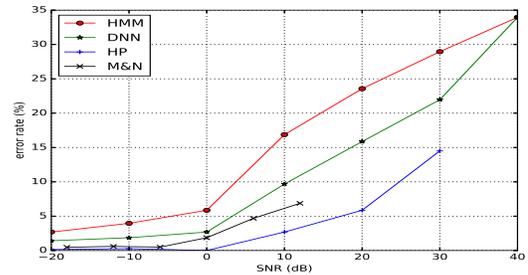}
\caption{Plot of distinctive-feature-distance results of human perception and HMM- and DNN- based ASR systems}
\label{fig:dfd}
\end{figure}

\subsection{Comparison of confusion matrices in white noise}
In order to examine error patterns in more detail, grey-scale confusion matrices for the detection results were constructed. The human perception results from the LAFF VCV database are shown in Fig. 5. As shown in the confusion matrices, in clean speech, most responses are on the diagonal, i.e. correct. In the next-to-last figure, the responses are spread out from the diagonal, indicating a more random detection pattern. In the final figure, all responses are given as "No Response", i.e. the human listeners were unable to answer. A similar pattern is seen for the Miller \& Nicely responses in Fig. 6, in that lower noise levels result in more diagonal entries, while higher noise levels result in a more random error pattern. However, the next-to-last confusion matrix shows that even at a low SNR of \textbf{-12dB}, the errors appear in blocks, which indicate common voicing features. For the confusion matrices for the HMM results shown in Fig. 7, the overall error rates are higher than for the human perception results, even at the higher SNR levels. Moreover, the results do not exhibit the characteristic blocking patterns that indicate similar voicing detections at noise levels around \textbf{-10dB} SNR. And finally, it can be seen that around \textbf{0dB} SNR or lower, most responses are given as \textbf{/s/}, which diverges from the human perception results. For the DNN results in Fig. 8, we can see that results are more similar to human perception results. However, we do not observe the characteristic blocking effect, and the responses at the lowest SNRs include a majority of \textbf{/s/} responses, similar to the HMM results. In order to quantify these effects, the distinctive-feature-distance is found as the proportion of features of the detected phoneme that deviate from the reference phoneme. For example, if the total number of features is 24, and the distinctive features can take on values of \textbf{+}, \textbf{-}, or \textbf{unspecified}, the largest mismatch can be 48. The normalized distinctive-feature-distances of the confusion matrices above are plotted in Fig. 9, which represents the sub-phonemic recognition performance of the ASR systems in comparison with the human perception results.

\section{Conclusions}
\label{sec:conclusions}
This study presents an evaluation method for sub-phonemic analysis of the effects of additive noise on two types of ASR systems, in direct comparison with human perception results.  Results were compared in terms of manner, place and voicing error patterns, as grey-scale confusion matrices, and as distinctive-feature-distances. Human perception results show that place features are most susceptible to misperception in white noise, followed by manner features, then voicing features. The DNN-based system showed similar patterns, although with more errors. In contrast, the HMM-based system had less consistency in the error patterns. In the confusion matrices, human perception results show that most errors occur near the diagonal regions, with blocking effects around -10dB SNR, which indicate correct voicing detections at that level. Meanwhile, in the DNN- and the HMM-based systems, errors converge to the sound \textbf{/s/} at higher noise levels, more randomized errors occur at higher SNR levels, and the blocking effect from voicing detection is not observed. The distinctive-feature-distances of the confusion matrices summarize the discrepancy of the ASR system performance from human perception results at the sub-phonemic level.

These results point to the possibility of incorporating parameters specifically related to voicing, manner, and place into acoustic models, and/or incorporating distinctive-feature-distance measures as training criteria for closer modeling of ASR systems to human perception patterns. Further work aims to extend this comparison to include results for other types of noise, e.g. babble and bandpass noise, which can be expected to produce perturbation patterns that are different from that from white noise for human perception and ASR systems.
\section{Acknowledgements}
\label{sec:acknowledgements}
The authors would like to thank Dr. Rachel Theodore and Nicholas Monto at the University of Connecticut-Storrs for their expertise in conducting the human listening experiments.

\vfill\pagebreak

\bibliographystyle{IEEEbib}
\bibliography{strings,refs}

\end{document}